\documentclass[10pt,twocolumn,letterpaper]{article}

\usepackage{iccv}              % To produce the CAMERA-READY version
\usepackage{url}

\usepackage{tabularx}
\usepackage{makecell}
\usepackage{algorithm}
\usepackage[accsupp]{axessibility} 

\usepackage{multirow}

\usepackage{bbding}

\usepackage{array}
\usepackage{listings}
\usepackage{setspace}
\usepackage{algorithmic}
\usepackage{pifont}
\usepackage{colortbl}
\usepackage{graphicx} % 引入 graphicx 宏包
\usepackage{amssymb}
\usepackage{fontenc} % 确保支持 T1 编码

\usepackage{tikz}
\newcommand{\blacklabel}[1]{%
	\begin{tikzpicture}[]
		\node[draw=none, fill=black, text=white, inner sep=0.35pt, font=\small, circle]{#1};
	\end{tikzpicture}
}

\definecolor{blue}{rgb}{0.0, 0.0, 1.0}

\definecolor{iccvblue}{rgb}{0.21,0.49,0.74}
\usepackage[pagebackref,breaklinks,colorlinks,allcolors=iccvblue]{hyperref}

\def\paperID{12288} % *** Enter the Paper ID here
\def\confName{ICCV}
\def\confYear{2025}

\title{Reminiscence Attack on Residuals: Exploiting Approximate Machine Unlearning for Privacy}

\vspace{-0.2cm}
\author{
	Yaxin Xiao$^\S$\\
	{\vspace{-0.085cm}\tt\small 20034165r@connect.polyu.hk\vspace{-0.05cm}}
	\and
	Qingqing Ye$^\S$\\
	{\vspace{-0.085cm}\tt\small qqing.ye@polyu.edu.hk\vspace{-0.05cm}}
	\and
	Li Hu$^\S$\\
	{\vspace{-0.085cm}\tt\small lily23.hu@polyu.edu.hk\vspace{-0.05cm}}
	\vspace{-0.3cm}
	\and
	Huadi Zheng$^\dagger$\\
	{\vspace{-0.085cm}\tt\small zhenghuadi@huawei.com\vspace{-0.05cm}}
	\and
	Haibo Hu$^{\S\ddagger*}$\\
	{\vspace{-0.085cm}\tt\small haibo.hu@polyu.edu.hk\vspace{-0.05cm}}
	\and
	Zi Liang$^{\S}$\\
	{\vspace{-0.085cm}\tt\small zi1415926.liang@connect.polyu.hk\vspace{-0.05cm}}
	\vspace{-0.3cm}
	\and
	Haoyang Li$^{\S}$\\
	{\vspace{-0.095cm}\tt\small hao-yang9905.li@connect.polyu.hk \vspace{-0.1cm}}
	\and
	Yijie Jiao$^{\S}$\\
	{\vspace{-0.095cm}\tt\small 22103037d@connect.polyu.hk \vspace{-0.1cm}}
	\and
	{\vspace{-0.3cm}\small $^{\S}$Department of Electronical and Electronic Engineering, The Hong Kong Polytechnic University ~$^\dagger$Huawei Technologies Co., Ltd. \vspace{-0.3cm}}\\
	\and
	{\small \vspace{-0.35cm}~$^\ddagger$Research Centre for Privacy and Security Technologies in Future Smart Systems ~$^*$Corresponding author: haibo.hu@polyu.edu.hk}
}

\begin{document}
\maketitle
\begin{abstract}
	Machine unlearning enables the removal of specific data from ML models to uphold the \textbf{ right to be forgotten}. While approximate unlearning algorithms offer efficient alternatives to full retraining, this work reveals that they fail to adequately protect the privacy of unlearned data. In particular, these algorithms introduce implicit residuals which facilitate privacy attacks targeting at unlearned data. We observe that these residuals persist regardless of model architectures, parameters, and unlearning algorithms, exposing a new attack surface beyond conventional output-based leakage. Based on this insight, we propose the \textbf{Reminiscence Attack (ReA)}, which amplifies the correlation between residuals and membership privacy through targeted fine-tuning processes. ReA achieves up to $1.90\times$  and $1.12\times$ higher accuracy than prior attacks when inferring class-wise and sample-wise membership, respectively. To mitigate such residual-induced privacy risk, we develop a dual-phase approximate unlearning framework that first eliminates deep-layer unlearned data traces and then enforces convergence stability to prevent models from ``pseudo-convergence'', where their outputs are similar to retrained models but still preserve unlearned residuals. Our framework works for both classification and generation tasks. Experimental evaluations confirm that our approach maintains high unlearning efficacy, while reducing the adaptive privacy attack accuracy to nearly random guess, at the computational cost of $2-12\%$ of full retraining from scratch.	
\end{abstract}

\vspace{-0.35cm}
\section{Introduction}
	\vspace{-0.1cm}
\label{sec:introduction}
As the significance of individual privacy grows, numerous laws and regulations have been established to grant individuals the right to protect their privacy. Notably, the General Data Protection Regulation (GDPR)~\cite{mantelero2013eu} and the Personal Information Protection and Electronic Documents Act (PIPEDA)~\cite{billtext} incorporate the \emph{right to be forgotten}, requiring organizations to delete user data upon request. In addition to traditional database systems, machine learning (ML) systems, which are trained on vast amounts of private data, must comply with this right by removing data from their training sets. This is critical because ML systems are vulnerable to data privacy breaches, such as membership inference attacks~\cite{mi17reza,xiao2022mexmi} and model inversion attacks~\cite{li2025sample}. 

The techniques that support such data removal in ML systems are collectively called machine unlearning (MU)~\cite{mu21lucas}. State-of-the-art (SOTA) machine unlearning methods can be categorized into exact unlearning and approximate unlearning. Exact unlearning necessitates retraining the model from scratch using only the remaining dataset, which is resource-intensive even with optimization techniques~\cite{mu21lucas}. In contrast, approximate machine unlearning (AMU) modifies the parameters of pre-trained machine learning models to approximate the retained model, allowing selective forgetting while maintaining efficiency. Currently, various AMU methods~\cite{2eternal20golatkar, 1modelsparisty23jia, 3machineunlearning23warnecke, 6can23chundawat,zhao2024what,huang2024unified} have achieved near-retrained unlearning efficacy when measured by legacy metrics such as prediction accuracy~\cite{1modelsparisty23jia}. Due to their significantly lower resource demands, AMU algorithms have become a focal point in this field.
%The unlearning efficacy of AMU is typically assessed by measuring the alignment between unlearned and retained models on accuracy across test, unlearned, and retained data, which are widely regarded as legacy metrics~\cite{1modelsparisty23jia}.

However, while machine unlearning initially aims to protect the privacy of the target data, it may inadvertently introduce new privacy risks. Such risks come from the imprint of unlearned data left behind in unlearned models, which becomes a source of privacy leakage. According to the imprint types, existing privacy attacks against unlearned data can be divided into two categories: version-dependent and version-independent attacks. Version-dependent privacy attacks exploit updates between historical model versions to infer the privacy of unlearned data~\cite{when21chen,hu2024learn,bertran2025reconstruction}. Thus, they are applicable only in scenarios where historical model versions are released. The second type, version-independent attacks, applies to broader scenarios and hence serves as the focus of this study. These attacks~\cite{kurmanji2023towards} detect privacy leaks by analyzing differences in the model's behavior on unlearned data versus non-training data. Yet, we argue that current version-independent  methods~\cite{carlini2022membership,kurmanji2023towards} only partially exploit the problem.

This paper investigates a critical oversight in version-independent privacy leakage analysis: Do existing approximate machine unlearning (AMU) methods inadvertently leak unlearned data privacy through \textbf{implicit residuals beyond the unlearned model's output distributions}? Prior works~\cite{carlini2022membership,kurmanji2023towards} focus on output-space residuals yet fail to capture the full extent of privacy risks. We reveal that AMU methods overlook latent residuals that alter their loss landscapes, leading to even greater risks for the membership privacy of unlearned data. To exploit these residuals, we introduce the Reminiscence Attack (ReA), which leverages a targeted fine-tuning approach to detect whether specific classes or samples were part of the training set.

\begin{figure}
	\centering
	\centering
	\subfloat[Class-wise Unlearning]{
		\includegraphics[width=0.41\columnwidth]{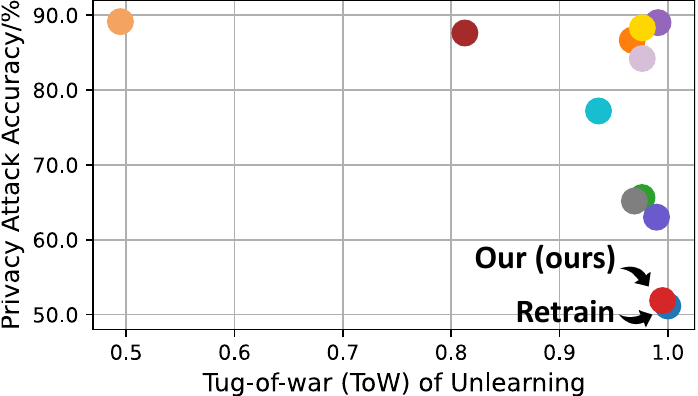}}~~~~
	\subfloat[Sample-wise Unlearning]{
		\includegraphics[width=0.41\columnwidth]{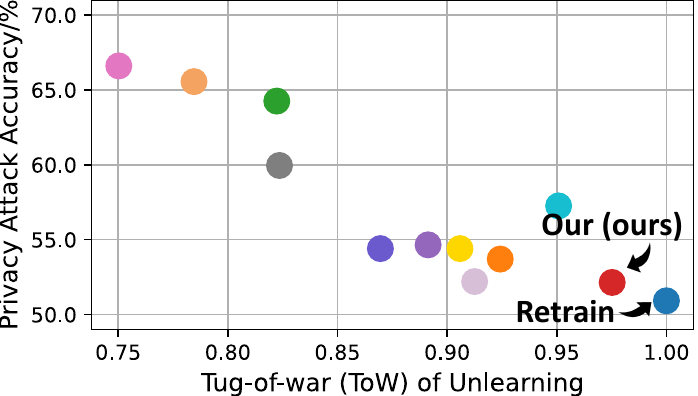}}\\
	\vspace{-0.08cm}
	\includegraphics[width=0.88\columnwidth]{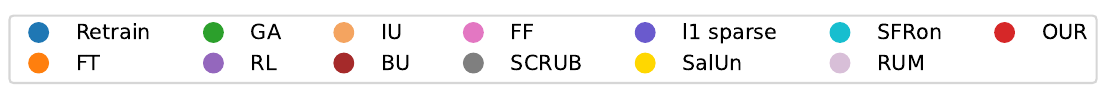}
	\vspace{-0.15cm}
	\caption{Comparison of unlearning efficacy (Tug-of-War metric, Equation~\ref{eq:tow}) and ReA attack accuracy across 12 unlearning methods tested in Section~\ref{subsec:ex_overall} on CIFAR20 (11 baselines and our OUR). OUR method approaches the ``Retrain'' in the down-right corner, showing high unlearning efficacy and low privacy leakage.}
	\vspace{-0.3cm}
	\label{fig:consistency_vs_privacy}
\end{figure}

To address such privacy leakages, we propose Orthogonal Unlearning \& Replay (OUR), an AMU method designed to scrub latent residuals in the model. OUR is built on two key insights: First, class-wise residuals persist because the unlearned model retains high intra-class correlation in hidden representations for the unlearned class, even when its predictions resemble those of retrained models. For deeper scrubbing, we enforce orthogonality between the hidden representations of unlearned data and their original values. Second, sample-wise residuals arise because the unlearned model stays in a ``pseudo-converged'' states that mimic the retrained model outputs (e.g., the prediction accuracy) but preserves latent information of unlearned data. To counteract this, we separate OUR into two phases to ensure model convergence through a final fine-tuning phase. In summary, our contributions are as follows:

\begin{itemize}
	\item  We introduce \emph{Reminiscence Attack} (ReA), a novel version-independent attack which targets the membership privacy of unlearned samples and classes. ReA uncovers privacy leakage beyond what existing attacks~\cite{kurmanji2023towards,1modelsparisty23jia} detect by exploiting residuals in the model's loss landscape.
	\item To address the privacy vulnerabilities exposed by ReA, we propose an approximate unlearning algorithm, Over-Unlearning \& Replay (OUR), designed to eliminate latent residuals within the unlearned models.
	\item Extensive experiments evaluate the unlearning efficacy of approximate unlearning algorithms and their privacy leakage using our ReA attacks, with results in classification scenarios summarized in Figure~\ref{fig:consistency_vs_privacy}. ReA effectively exposes unlearned data privacy, achieving up to $90\%$ attack accuracy for unlearned class membership inference. The results also confirm that OUR significantly reduces ReA's attack accuracy to near-random levels while preserving high unlearning efficacy and efficiency.
	
\end{itemize}

%[TODO]The paper is organized as follows. Section~\ref{sec:preliminary} formalizes key concepts and threat models, Section~\ref{sec:attack} develops the \emph{Reminiscence Attack} (ReA), and Section~\ref{sec:optimized_unlearning} introduces our Orthogonal Unlearning \& Replay (OUR) framework. Section~\ref{sec:examination} empirically evaluates ReA's effectiveness in exposing MU privacy vulnerabilities and OOUR'sdefense capabilities. Section~\ref{sec:related_works} contextualizes our contributions within existing literature, and Section~\ref{sec:conclusion} concludes the paper.

\section{Problem Definition}
\label{sec:preliminary}
\subsection{Machine Unlearning (MU)}
Machine unlearning  (MU) removes the influence of specific training data on a pre-trained machine learning model $F(\cdot;\theta)$ through exact or approximate methods. While exact unlearning requires retraining from scratch, approximate unlearning modifies model parameters $\theta$ to achieve comparable unlearning efficacy at significantly lower computational cost. Formally, given a model $F(\cdot;\theta)$ trained on dataset $\mathcal{D}$, let $\mathcal{D}_u$ denote unlearned data and $\mathcal{D}_r = \mathcal{D} \setminus \mathcal{D}_u$ to the retained data. The approximate unlearning algorithm $\mathcal{U}$ aims to produce parameters $\theta^u = \mathcal{U}(\theta)$ such that $F(\cdot;\theta^u)$ approximates the model $F(\cdot;\theta^0)$ retrained on $\mathcal{D}_r$.

\noindent\textbf{Unlearning in Classification}. In classification tasks, unlearning operates at two granularities: \textbf{sample-wise unlearning} for individual data points and \textbf{class-wise unlearning} for entire classes~\cite{2eternal20golatkar}. Specifically, sample-wise unlearning processes $\mathcal{D}_u$ as discrete samples, requiring the unlearned model $F(\cdot;\theta^u)$ to match the retrained model $F(\cdot;\theta^0)$'s behavior by treating $\mathcal{D}_u$ as unseen test data. Class-wise unlearning handles $\mathcal{D}_u$ as entire classes, and demands $F(\cdot;\theta^u)$ aligns with $F(\cdot;\theta^0)$ to recognize unlearned classes as out-of-distribution (OOD) classes. This ensures the unlearned model has identical task scopes to $F(\cdot;\theta^0)$.

\noindent\textbf{Unlearning in Image Generation}. For generative models such as text-to-image diffusion models~\cite{rombach2022high}, we shift focus from safety alignment to privacy preservation. While existing methods focus on removing harmful contents (e.g., nudity)~\cite{fan2024salun,gao2024meta}, we address \textbf{identity-specific concept unlearning} in portrait generation. This process aims to eliminate identity-specific concepts from $F(\cdot,\theta)$ while maintaining output quality for retained tasks.

\noindent\textbf{Legacy MU Metrics}. Machine unlearning is assessed with the following legacy metrics~\cite{1modelsparisty23jia,3machineunlearning23warnecke,li25funu}: \blacklabel{1}{\bf Test Accuracy (TA)}. It is defined as TA$=Acc_{\mathcal{D}_t}(F(.;\theta^u))$, which measures the accuracy of $F(.;\theta^u)$ on the test dataset $\mathcal{D}_t$. \blacklabel{2}{\bf Unlearning Accuracy (UA)}. It is calculated by UA$=Acc_{\mathcal{D}_u}(F(.;\theta^u))$. \blacklabel{3}{\bf Retaining Accuracy (RA)}. It's defined as RA$=Acc_{\mathcal{D}_r}(F(.;\theta^u))$. The closer the three accuracy metrics are to the `ground-truth' retrained model $F(.;\theta^0)$, the better the unlearning efficacy. \blacklabel{4}{\bf Membership Inference Attack on} $\mathcal{D}_u$ {\bf (MIA Efficacy)}. This metric quantifies the success rate of membership inference attacks~\cite{mi17reza} in detecting $\mathcal{D}_u$ as training data for $F(\cdot;\theta^u)$. Unlike prior evaluations~\cite{1modelsparisty23jia} measuring only the true positive rate (TPR), we adopt TPR at a fixed 0.1 false positive rate (FPR) for stricter privacy assessment. \blacklabel{5}{\bf Run-Time Efficacy (RTE)}. This metric refers to the time a machine unlearning algorithm consumes, which assesses its practical efficiency. \blacklabel{6}\textbf{Tug-of-War (ToW)}~\cite{zhao2024what}. This unified metric quantifies the alignment between unlearned ($\theta^u$) and fully retrained ($\theta^0$) models across multiple metrics $M_1$, $M_2$, and $M_3$. When instantiated with TA, UA, and RA, ToW measures unlearning efficacy. Let $d(\cdot,\cdot)$ denote the absolute distance between metric values. The ToW score is defined as:
\vspace{-0.27cm}
\begin{equation}
	\footnotesize
	\label{eq:tow}
	\text{ToW}(M_1, M_2, M_3) = \prod\nolimits_{i=1}^3(1 - \frac{d(M_i(\theta^u), M_i(\theta^0))}{M_i(\theta^0)}).
	\vspace{-0.07cm}
\end{equation}

\subsection{Privacy Attacks Targeting MU} 
\label{subsec:privacy_attacks}
Privacy attacks compromise unlearned data privacy via exploring information leakage in machine unlearning processes. These attacks fall into two categories: (1) version-dependent attacks, which analyze model update histories~\cite{when21chen,hu2024learn,bertran2025reconstruction}, and (2) version-independent attacks, which leverage residuals of approximate unlearning. As version management addresses only the former, we focus on the latter category which exposes vulnerabilities in approximate unlearning methods. This category includes two benchmarks: (1) \textbf{MIA-LiRA}. It applies SOTA membership inference attacks (MIA) LiRA~\cite{carlini2022membership,bai25membership}, which uses shadow models and a likelihood ratio test on model confidence to detect unlearned data. Since its attack model is designed to detect training data for the original model $F(\cdot;\theta)$, its effectiveness diminishes when unlearned and training data yield distinct output distributions. (2) \textbf{MIA against User Privacy (MIA-UP)}~\cite{kurmanji2023towards}. This attack distinguishes unlearned data from test data. While adopting LiRA's shadow model framework, this method uniquely synchronizes shadow models with the unlearned model's unlearning process. Then, the binary classifier attacker trains on the outputs of shadow models, using unlearned data as positives and test data as negatives. However, it becomes unstable if shadow and unlearned models' unlearning processes diverge. Up to now, privacy attacks targeting unlearned data primarily exploit residuals in output distributions.

\subsection{Threat Model}
\label{sub:tm}
%While existing privacy attacks on machine unlearning focus on sample-level membership inference, we pioneer the formalization of \textbf{class membership privacy} leaks, which infer whether a specific class has ever been unlearned. We assume adversaries are aware of the task scope and have either white-box access or black-box access restricted to API queries.

While existing privacy attacks on machine unlearning focus on sample-level membership inference, we pioneer the formalization of \textbf{class membership privacy} leaks, which infer whether specific classes were ever unlearned. We assume adversaries are aware of the task scope and have either white-box or black-box API access.

\noindent\textbf{Sample-wise Membership Inference Attack (MIA)}. For models undergoing sample-wise unlearning ($F(·;\theta^u$)), adversaries infer membership of individual unlearned samples, where test data is negative references. Following SOTA MIA evaluations~\cite{carlini2022membership,bai2025rmr}, we analyze the true-positive rate (TPR) versus the false-positive rate (FPR) of attacks (i.e., the precision-recall tradeoff) rather than fixed accuracy. Specifically, the attack model outputs continous confidence scores through $\mathcal{A}^\prime$, which are then thresholded by $\tau$ to yeild a membership prediction. Formally, for an inferred sample $x$ with label $y$, the output of the sample-MIA attacker $\mathcal{A}_\text{sample}$ is: 
\vspace{-0.1cm}
\begin{equation}
	\footnotesize
	\label{eq:sample_mia}
	\mathcal{A}_\text{sample}(x, y) = \mathbf{1}[\mathcal{A}^{\prime}(F(x;\theta^u),y)>\tau],
	\vspace{-0.1cm}
\end{equation}
where $\mathbf{1}$ is the indicator function and $\tau$ controls the precision-recall tradeoff.

\noindent\textbf{Class-wise Membership Inference Attack (MIA)}. For models undergoing  $F(.;\theta^u)$ undergoes class-wise unlearning, the adversary infers membership of candidate OOD classes $\mathcal{D}^\text{ood}$, i.e., determining whether it was unlearned. Crucially, this attack can extend to concept unlearning in generative tasks, targeting the membership of forgotten concepts. Mirroring the sample-wise formulation (Equation~\ref{eq:sample_mia}), the output of the class-wise MIA attacker $\mathcal{A}_\text{class}$ for  $\mathcal{D}^\text{ood}$ is:
\vspace{-0.08cm}
\begin{equation}
	\footnotesize
	\label{eq:class_mia}
	\mathcal{A}_\text{class}(\mathcal{D}^\text{ood}) =\mathbf{1}[\mathcal{A}^{\prime}(\mathcal{D}^\text{ood})>\tau].
	\vspace{-0.2cm}
\end{equation}

\vspace{-0.4cm}
\section{Reminisence Attack (ReA): Exploiting Loss Landscape Residuals}
\label{sec:attack}
\vspace{-0.1cm}
We propose the Reminiscence Attack (ReA), a novel version-independent membership inference attack against unlearned samples, classes, and concepts. This method exploits latent residuals preserved in the loss landscape during approximate unlearning, establishing an orthogonal attack surface to existing approaches (MIA-LiRA/UP).

\begin{figure}
	\centering
	\hspace{-0.3cm}\subfloat[Class-wise Unlearn (RUM~\cite{zhao2024what})]{
		\label{fig:class_wise_landscape}
		\includegraphics[width=0.468\columnwidth]{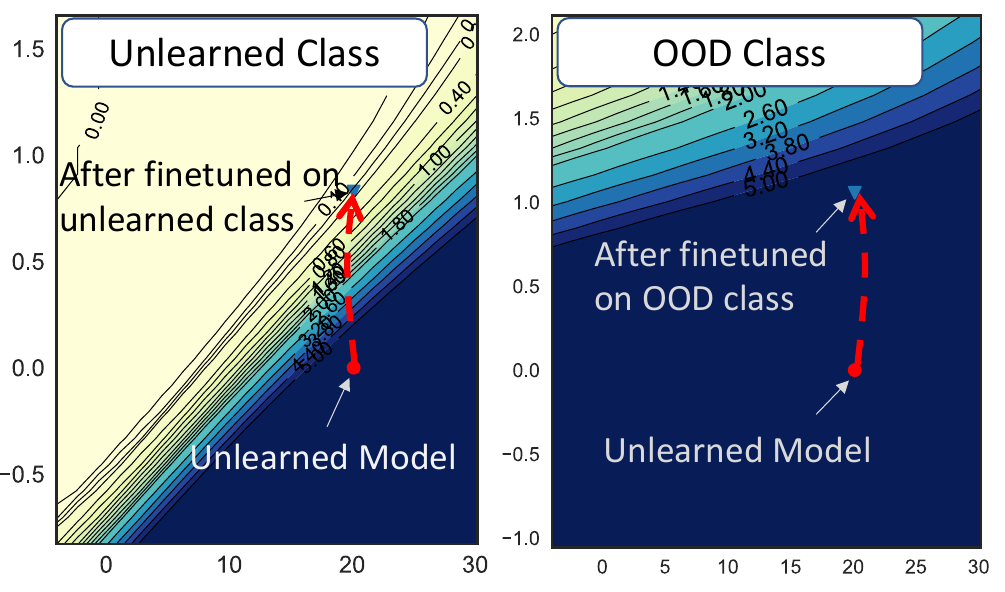}}
        \subfloat[Sample-wise Unlearn (RL~\cite{amnesiac21graves})]{
		\label{fig:sample_wise_landscape}
		\includegraphics[width=0.468\columnwidth]{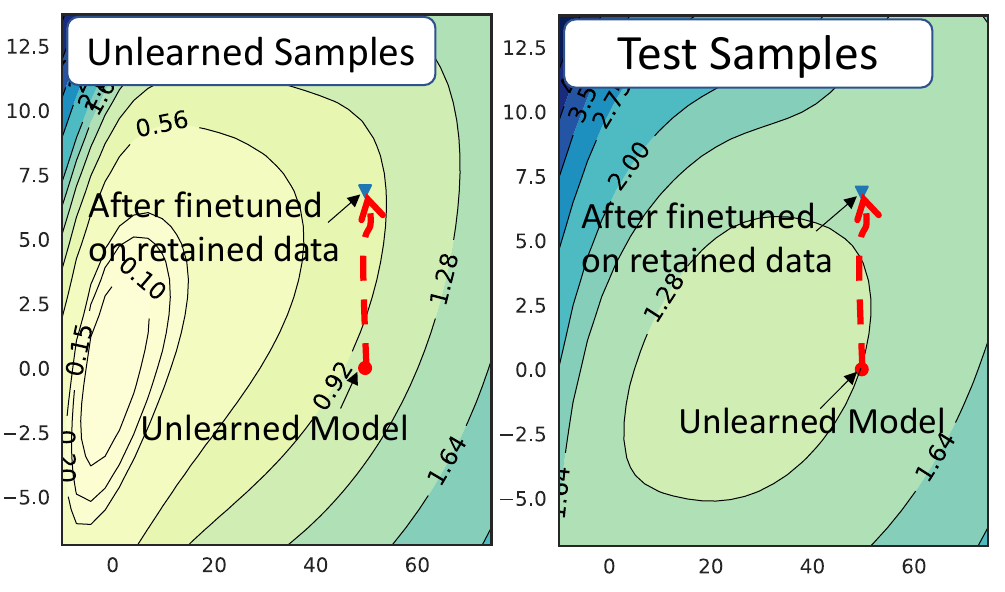}}
	\vspace{-0.25cm}
	\caption{Visualized loss landscapes around unlearned models in CIFAR20 experiments. Dashed lines are guided fine-tuning trajectories. Appendix~\ref{app:loss_landscape} provides additional results on other datasets.}%, visualized by the method in~\cite{zheng2025spurious}.}

\vspace{-0.35cm}
\label{fig:landscape}
\end{figure}

\noindent\textbf{Understanding Unlearning Residuals in Loss Landscapes}. Approximate unlearning may produce measurable residuals that deviate from `gold-standard' retrained models, causing persistent privacy risks. Formally, for an unlearned model $F(\cdot;\theta^u)$ obtained by unlearning dataset $\mathcal{D}_u$, we define its residual as 
\begin{equation}
	\footnotesize
	R(\mathcal{D}_u):=\mathbb{E}\left[d(F(\mathcal{D}_u;\theta^u), F(\mathcal{D}_\text{non-tr};\theta^u))\right],
\end{equation}		 
where $\mathcal{D}_\text{non-tr}$ is the non-training counterpart to $\mathcal{D}_u$. $d$ is a task-specific metric measuring model behavior discrepancy. We reveal that these residuals not only manifest in output distributions but also reshape the loss landscape. As visualized in Figure~\ref{fig:class_wise_landscape}, unlearned models remain near optima (bright regions) for $\mathcal{D}_u$ data and converge rapidly under guided finetuning, revealing latent traces, contrasting with non-training data (including OOD classes and test samples). %we visualize the loss landscapes of unlearned data versus non-training data (including OOD classes and test samples). The unlearned models stay near optimal solutions (bright regions) for unlearned data, revealing its latent traces.

To efficiently expose membership information in the loss landscape, targeted fine-tuning is the natural choice, which we term \textbf{reminiscence}. Its implementation depends on the inference target. In class-wise unlearning, as shown in Figure~\ref{fig:class_wise_landscape} (left), the unlearned model lies on flat plateaus near sharp minima. Guiding fine-tuning toward the minima accelerates convergence for unlearned classes while slowing non-training classes, which reveals membership. This guidance is achieved by introducing retained data in fine-tuning, as the targeted path aligns with their optimal solutions.

In sample-wise unlearning, as shown in Figure~\ref{fig:sample_wise_landscape} (left), the unlearned model remains on a steep loss landscape for unlearned data, indicating incomplete convergence. If the model shifts to the low-loss basin (bright regions), the loss gap between unlearned and test data widens, further exposing membership. Unlike class-wise ReA, inferred samples do not share memberships and cannot be fine-tuned in groups. However, fine-tuning on retained samples still guides the model to the low-loss basin, as their optimal solution regions align with unlearned samples. While both scenarios rely on retained samples, Section~\ref{subsec:rml} explains how to remove this assumption.

%\begin{figure}
%	\centering
%	\hspace{-0.2cm}\subfloat[Class-wise unlearning (RUM~\cite{zhao2024what})]{
%		\label{fig:class_wise_landscape}
%		\includegraphics[width=0.93\columnwidth]{figure/landscape/landscape_class1.pdf}}\\
%	\hspace{-0.2cm}\subfloat[Sample-wise unlearning (RL~\cite{amnesiac21graves})]{
%		\label{fig:sample_wise_landscape}
%		\includegraphics[width=0.93\columnwidth]{figure/landscape/landscape_sample_rl.pdf}}
%	\vspace{-0.25cm}
%	\caption{Visualized loss landscapes of unlearned data and non-training data around unlearned models in CIFAR20 experiments.}%, visualized by the method in~\cite{zheng2025spurious}.}
%	\vspace{-0.35cm}
%	\label{fig:landscape}
%\end{figure}

\subsection{Reminiscence Attack (ReA)}
\label{subsec:rml}
Earlier, we introduced the method of reminiscence (targeted fine-tuning) to leverage residuals in the loss landscape to amplify the membership leakage of unlearned data. Here, we detail the steps of the reminiscence attack (ReA).

\noindent\textbf{Class-wise ReA}. To detect the link between residuals and an unlearned class, ReA reintroduces the inferred class (a set $\mathcal{D}^\text{ood}$) into training and measures how quickly it resonates with residuals to derive the attack confidence score. Formally, this is quantified by the resonance index $\text{Idx}\text{r}$, defined as the number of iterations required for the model to achieve high prediction accuracy (i.e., convergence threshold) on $\mathcal{D}^\text{ood}$. Then, the confidence score in Equation~\ref{eq:class_mia} is computed as: 
\vspace{-0.1cm}
\begin{equation} 
	\label{eq:conf_class} \footnotesize \mathcal{A}^\prime(\mathcal{D}^\text{ood})=1-\text{Idx}_\text{r}(\mathcal{D}^\text{ood})/\text{Idx}_\text{max},
	\vspace{-0.1cm}
\end{equation} 
where $\text{Idx}_\text{max}$ is the total number of training iterations. A lower $\text{Idx}\text{r}$ means faster convergence and suggests the inferred class closer to the unlearned class in Figure~\ref{fig:class_wise_landscape}, which leads to a higher confidence score in Equation~\ref{eq:conf_class} when inferred as unlearned data. The above method assumes a white-box attacker, but this assumption can be relaxed with substitute models from model extraction~\cite{truong2021data}, given the demonstrated transferability of residuals in Appendix~\ref{subsec:scenario}.

There are two key challenges affecting class-wise ReA's efficiency. First, ReA ideally requires the retained dataset to guide the shortest optimal path, but attackers may lack access to it. This issue can be mitigated by minimizing logits updates from OOD samples (excluding the inferred class)~\cite{lv2023usenix}, as this fulfills the retained dataset’s key role in maintaining utility. Second, the fine-tuned learning rate (lr) significantly impacts ReA's stability. To address this, we introduce the \textbf{multi-lr aggregation strategy}, which averages resonance indexes across multiple learning rates $\{\text{lr}_j\}_{j=1}^J$. Thus, the revised confidence score is:
\vspace{-0.05cm}
\begin{equation}
	\label{eq:aggre_conf_class}\footnotesize
	\mathcal{A}^\prime(\mathcal{D}^\text{ood})=1-\left[\sum\nolimits_{j=0}^J{\text{Idx}_\text{r}(\mathcal{D}^\text{ood}, \text{lr}_j})/(J\cdot \text{Idx}_\text{max})\right].
	\vspace{-0.07cm}
\end{equation}

\noindent\textbf{Sample-wise ReA}. In sample-wise unlearning, residuals persist primarily due to incomplete convergence. To fully expose the membership of unlearned samples, ReA fine-tunes the unlearned model on the retained dataset with a reduced learning rate for convergence. If residuals remain, the model’s accuracy on unlearned samples rises abnormally higher than on test data, signaling additional privacy leakage. Building on this insight, ReA follows two steps: (1) \textbf{the reminiscence process}, where the unlearned model is fine-tuned on the retained dataset, and (2) \textbf{the MIA process}, where membership is inferred using MIA-LiRA~\cite{carlini2022membership}. MIA-LiRA is preferred over MIA-UP to avoid dependence on unlearning algorithm knowledge. However, the first step requires knowledge of the retained data identity, which is often inaccessible to attackers. To address this, we propose inferring a set of high-confidence positive samples using MIA-LiRA, and them as a ``pseudo'' retained dataset in the first step. If white-box access is unavailable, sample-wise ReA reduces to MIA-LiRA. Appendix~\ref{app:algorithm} provides pseudo codes for class-wise and sample-wise ReA.

\vspace{-0.2cm}
\section{Scrubbing Residuals in Unlearning: Orthogonal Unlearning \& Replay (OUR)}
\label{sec:optimized_unlearning}
Previous analysis indicates that if residuals remain from approximate machine unlearning (AMU) processes, particularly in the overlooked loss landscape, they may unintentionally facilitate membership inference attacks (e.g., ReA) against unlearned samples. This highlights the necessity to develop AMU methods that not only align output distributions with retrained models but also scrub residuals, to ensure privacy protection for unlearned data.

\subsection{High-level Solution of Scrubbing Residuals}
To scrub residuals, we must understand their cause and address both class-wise and sample-wise cases. Class-wise residuals persist when the collective features of the unlearned class remain in deeper hidden layers (far from the output layer), as confirmed in Section~\ref{subsec:visual_reps}. To eliminate these traces, we enforce orthogonality between the hidden and original features of unlearned samples, achieving deeper scrubbing beyond existing methods (e.g., random label~\cite{amnesiac21graves}). This process is termed as orthogonal unlearning.

For sample-wise cases, residuals arise from incomplete convergence on retain tasks, where unlearned accuracy (UA) falsely resembles test samples' behavior. Resolving this requires the model to reach an actual convergence state. A single loss function optimizing both and model utility often hinders this process, so we structure unlearning in \textbf{two phases}:  (1) Orthogonal Unlearning to deeply scrub unlearned sample information, and (2) replay on the retained set to restore convergence and preserve utility. Our notion of orthogonality differs from prior orthogonal projection unlearning~\cite{wang2025precise,hoang2024learn,gao2024ethos}, which perturbs directions orthogonal to retaining tasks to preserve utility. Instead, we minimize the cosine similarity between unlearned features and their original state to scrub unlearning residuals.
%This two-phase framework, named Orthogonal Unlearning \& Replay (OUR), provides a principled approach to effectively scrubbing residuals.

\subsection{Orthogonal Unlearning \& Replay (OUR)}
Next, we detail the two phases of OUR. Let $F_l(:;\theta)$ represent the output of the $l$-th layer of the unlearned model (i.e., representations/feature), and $\theta^0$ be the pre-unlearning states. In the first phase, the model is trained over $e_1$ epochs with the following loss function:
\vspace{-0.15cm}
\begin{equation}
	\footnotesize
	L_\text{orth} = \sum\nolimits_{(x, y)\in\mathcal{D}_u} \sum\nolimits_{l\in\{l\}_k}[F_l^T(x; \theta)F_l(x; \theta^0)]^2.
	\vspace{-0.1cm}
\end{equation}

$L_\text{orth} $ encourages orthogonality between the the features of $F(\cdot; \theta)$ and $F(\cdot; \theta^0)$ on unlearned samples. Here, $\{l\}_k$ denotes the set of $k$ hidden layers selected for this phase.

During the replay phase, the unlearned model is fine-tuned on $D_r$ for $e_2$ epochs, yielding the final unlearned model $F(.;\theta^u)$. To enhance unlearning effectiveness, we incorporate $l1$ regularization optimization~\cite{1modelsparisty23jia}. The pseudo-code for OUR is provided in Appendix~\ref{app:algorithm}.

\noindent\textbf{Key Considerations of OUR}. 
How is run-time efficacy (RTE) maintained? RTE is ensured by excluding the retained set in the orthogonal unlearning phase. While this may induce immediate utility degradation, the replay phase achieves rapid recovery due to the minimal change of neurons. Our analysis in Appendix~\ref{app:our_stable_neurons} shows that most neurons undergo trivial changes during unlearning, allowing efficient recovery in the replay phase.

How to prevent model collapse during orthogonal unlearning? Model collapse occurs when parameter gradients explode. Therefore, the orthogonal unlearning phase should constrain the maximal parameter change $\Delta_\text{max}\leq\Delta_\text{thr}$. $\Delta_\text{thr}$ is a value (e.g., $5e-3$) which may trigger gradient exposure during replay and set as the maximum distance between two random models (see Appendix~\ref{app:our_stable_neurons}).

\section{Experimental Evaluation}
\label{sec:examination}
In this section, we first evaluate the privacy vulnerabilities of $11$ approximate unlearning benchmarks using \emph{Reminiscence Attack} (ReA). We then analyze the unlearning efficacy and privacy leakage of the proposed \emph{Orthogonal Unlearning \& Replay} (OUR) framework. Our code is available at \url{https://github.com/orthogonalunlearning-replay/OUR}.

\subsection{Setup}
\label{subsec:ex_setup}
\noindent\textbf{Datasets and Models}. We conduct classification experiments on three benchmarks: CIFAR10 with ResNet18, CIFAR100 with ResNet18~\cite{he16deep}, and CIFAR20 (20 broad classes aggregated from CIFAR100's fine-grained labels) with Vision Transformer (ViT)~\cite{dosovitskiy2021an}. %yoshioka2024visiontransformers
Class-wise unlearning is performed on CIFAR100 and CIFAR20, while sample-wise unlearning applies to CIFAR10 and CIFAR20. In the image-generation domain, we specialize in unlearning for text-to-image Stable Diffusion (SD)~\cite{rombach2022high} with a focus on privacy. Specifically, we perform identity-level unlearning on an SD model fine-tuned via DreamBooth~\cite{ruiz2023dreambooth} using CelebA-HQ~\cite{karras2018progressive} identities.

\noindent\textbf{Unlearning Configuration}. We benchmark 6 SOTA AMU algorithms and 5 optimized frameworks (Appendix~\ref{app:mu_benchmark}). For OUR, $\{l\}_k$ uses the first, middle, and last block outputs, as verified in Appendix~\ref{app:exp_abl_layerchoice}. Other detailed OUR setups are in Appendix~\ref{app:ex_setups_sub}. For class-wise unlearning, each trial selects five random classes as out-of-distribution (OOD) and one target class for unlearning. Using only one OOD class could introduce bias due to class-specific differences. Thus, corresponding experiments are labeled CIFAR20 (-5) and CIFAR100 (-5). MU evaluation metrics follow Section~\ref{sec:preliminary}.

\noindent\textbf{Privacy Attack Parameters}. Class-wise ReA adversaries have limited access to unlearned data, with only 3\% for CIFAR20 and 20\% for CIFAR100. Each attack experiment consists of 50 trials with randomized class selections, totaling 300 inferred classes. ReA uses an SGD optimizer and an lr set $\{0.001, 0.005, 0.007, 0.01\}$ to perform a multi-lr aggregration strategy. Its convergence threshold is set to $75\%$. The ablation studies discussing hyperparameters of ReA are presented in Appendix~\ref{subsec:rea_lr} and Appendix~\ref{app:rea_threshold}. For sample-wise MIA, following prior work~\cite{kurmanji2023towards}, we assume adversaries have a dataset containing training (retained), unlearned, and test data but lack knowledge of their memberships. Here, MIA specifically focuses on the membership of unlearned data rather than training (retained) data. Sample-wise ReA uses an lr of $0.01$ for CIFAR10 (SGD) and $5$e$^{-5}$ for CIFAR20 (Adam), which is $0.1\times$ the training lr. 

Evaluation for MIA follows SOTA ROC analysis~\cite{carlini2022membership}, measuring MIA performance with TPR at a fixed FPR. Balanced attack accuracy is also reported. Detailed descriptions of OUR hyper-parameters, ReA settings, and platforms are provided in Appendix~\ref{app:ex_setups_sub}.

\noindent\textbf{Privacy Attack Benchmarks}. To compare with ReA, we evaluate two SOTA attacks: MIA-LiRA~\cite{carlini2022membership} and MIA-UP~\cite{kurmanji2023towards}, as described in Section~\ref{sec:preliminary}. Though originally designed for sample-wise MIA, we adapt them to class-wise MIA as follows. Both methods run on the entire set of inferred classes which match the training set size, producing a single positive rate with a fixed threshold $\tau'$. This score serves as the output of $\mathcal{A}^\prime$ in Equation~\ref{eq:class_mia}. Specifically, MIA-UP replaces shadow models' test samples with OOD classes to serve as the negative samples. MIA-LiRA assumes knowledge of model architecture and hyper-parameters, while MIA-UP further requires unlearning algorithm details.

\subsection{Evaluation Results in Classification Tasks}
\label{subsec:ex_overall}
We present class-wise and sample-wise unlearning performance for classification tasks in Tables~\ref{tb:mu}, with subscripts indicating standard deviations and blue values showing differences from the retrained model. The corresponding class-wise and sample-wise privacy attack results for different AMU methods are shown in Table~\ref{tb:overall_evaluation_class} and Table~\ref{tb:overall_evaluation_sample}, with bold denoting the best results. Table~\ref{tb:mu} demonstrates that our OUR achieves unlearning efficacy comparable to the ground-truth retrained model on both ViT and ResNet architectures, while exhibiting superior run-time efficiency (RTE) which is lower than $8\%$ of retraining time. %Notably, class-wise influence unlearning (IU)~\cite{1modelsparisty23jia} and boundary-type unlearning (BU)~\cite{chen2023boundary} show limited effectiveness on ViT models, with minimal class removal capability.

Besides, the privacy vulnerability evaluation reveals two critical insights (Tables~\ref{tb:overall_evaluation_class}-\ref{tb:overall_evaluation_sample}). First, our Reminiscence Attack (ReA) significantly outperforms existing privacy attacks (MIA-LiRA and MIA-UP) across most MU methods. Class-wise MU methods such as random label (RL)~\cite{amnesiac21graves}, and its optimized version Salun~\cite{fan2024salun} and RUM~\cite{zhao2024what}, IU~\cite{1modelsparisty23jia}, and BU~\cite{chen2023boundary} exhibit severe vulnerabilities, with ReA achieving over $81\%$ accuracy. This vulnerability likely arises because RL-based methods mainly affect model parameters near the output layer, and IU and BU have limited unlearning efficacy on ViT models. For sample-wise unlearning, IU~\cite{1modelsparisty23jia} and FF~\cite{2eternal20golatkar} which lack gradient descent processes, and gradient ascent (GA)~\cite{thudi2022unrolling} show heightened susceptibility to the reminiscence process, with ReA attaining over $55\%$ (on CIFAR10) and $70\%$ (on CIFAR20) accuracy. Second, our OUR method substantially mitigates privacy leakage and reduces class-wise and sample-wise MIA accuracies lower than $54.73\%$ and $53.80\%$, respectively, which are significantly close to retraining baselines.

\begin{figure}
	\centering
	\subfloat[Retrain]{
		\includegraphics[width=0.29\columnwidth]{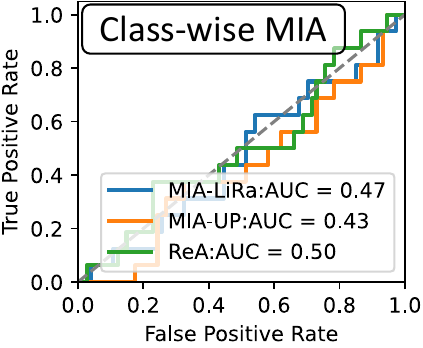}}~
	\subfloat[RUM~\cite{zhao2024what}]{
		\includegraphics[width=0.29\columnwidth]{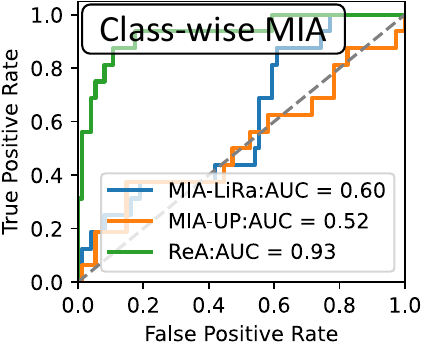}}~
	\subfloat[OUR]{
		\includegraphics[width=0.29\columnwidth]{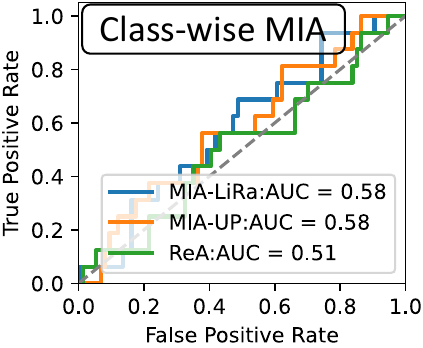}}\\
	\subfloat[Retrain]{
		\includegraphics[width=0.29\columnwidth]{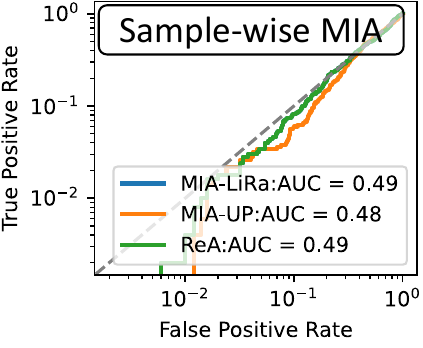}}~
	\subfloat[SCRUB~\cite{kurmanji2023towards}]{
		\includegraphics[width=0.29\columnwidth]{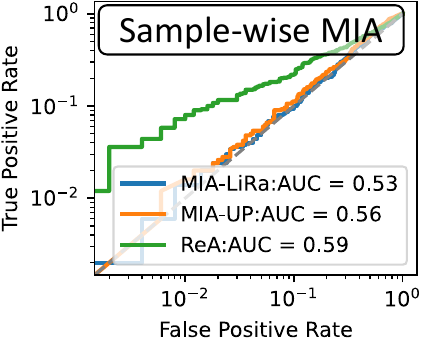}}~
	\subfloat[OUR]{
		\includegraphics[width=0.29\columnwidth]{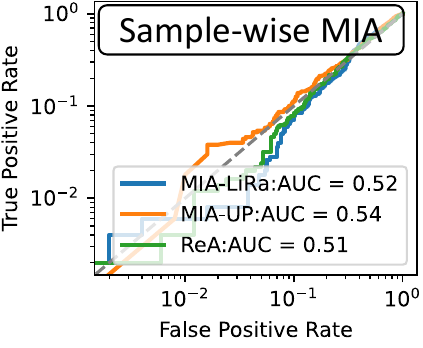}}
	\vspace{-0.31cm}
	\caption{ROC Analysis of MIA (Selected Methods). }%Appendix~\ref{app:roc_full} provides ROC analysis for all AMU methods.}
	\vspace{-0.575cm}
	\label{fig:auc_class_wise}
\end{figure}

\noindent\textbf{ROC Analysis of Privacy Attacks}. ROC analysis~\cite{carlini2022membership} MIA via the trade-off between true positive rate (TPR) and false positive rate (FPR). The area under the curve (AUC) quantifies attack effectiveness, with higher values indicating better separation between member and non-member distributions. Figures~\ref{fig:auc_class_wise} compare ROC curves for MU benchmarks, OUR, and retrained models on CIFAR20 in class-wise MIA and sample-wise MIA, respectively. The results indicate that ReA achieves substantially higher AUC values (up to $1.78\times/1.55\times$) than MIA-LiRA/UP, demonstrating superior membership inference capability. Notably, OUR's curves align closely with retrained models (AUC difference $<0.11$), suggesting better privacy preservation. Complete ROC analyses of all AMU appear in Appendix~\ref{app:roc_full}.

\begin{table}
	\caption{Machine Unlearning (MU) for Classification}
	\vspace{-0.28cm}
	\label{tb:mu}
	\tiny
	\begin{spacing}{1.0}
		%\begin{tabular}{|p{1.15cm}<{\centering}|p{0.875cm}<{\centering}|p{0.905cm}<{\centering}|p{0.905cm}<{\centering}|p{0.905cm}<{\centering}|p{0.905cm}<{\centering}|}
		\begin{tabular}{p{1.01cm}<{\centering}p{1.0cm}<{\centering}p{1.0cm}<{\centering}p{1.0cm}<{\centering}p{1.25cm}<{\centering}p{0.56cm}<{\centering}}
			\hline
			\multicolumn{6}{c}{\textbf{Class-wise Unlearning}}\\ \hline
			\makecell{MU}&\makecell{TA/\%}&\makecell{UA/\%}&\makecell{RA/\%}&\makecell{MIA Efficacy/\%} &\makecell{RTE (s)}\\ \hline
			\rowcolor{gray!7}
			\multicolumn{6}{c}{CIFAR20 (-5) - ViT}\\ \hline
			Retrain &51.31\scalebox{0.7}{$\pm$0.11}(\textcolor{blue}{0.00})&0.00\scalebox{0.7}{$\pm$0.00}(\textcolor{blue}{0.00})&99.99\scalebox{0.7}{$\pm$0.01}(\textcolor{blue}{0.00})&1.67\scalebox{0.7}{$\pm$0.10}(\textcolor{blue}{0.00})&1801.26\\
			FT~\cite{amnesiac21graves}            &49.40\scalebox{0.7}{$\pm$0.27}(\textcolor{blue}{1.91})&0.74\scalebox{0.7}{$\pm$0.58}(\textcolor{blue}{0.74})&97.89\scalebox{0.7}{$\pm$0.04}(\textcolor{blue}{2.10})&1.54\scalebox{0.7}{$\pm$0.18}(\textcolor{blue}{0.13})&120.99\\
			GA~\cite{thudi2022unrolling}            &49.13\scalebox{0.7}{$\pm$0.55}(\textcolor{blue}{2.18})&0.00\scalebox{0.7}{$\pm$0.00}(\textcolor{blue}{0.00})&99.01\scalebox{0.7}{$\pm$0.02}(\textcolor{blue}{0.98})&92.05\scalebox{0.7}{$\pm$3.13}(\textcolor{blue}{90.38})&69.46\\
			RL~\cite{amnesiac21graves}       &50.72\scalebox{0.7}{$\pm$0.15}(\textcolor{blue}{0.59})&0.00\scalebox{0.7}{$\pm$0.00}(\textcolor{blue}{0.00})&99.98\scalebox{0.7}{$\pm$0.01}(\textcolor{blue}{0.01})&1.66\scalebox{0.7}{$\pm$0.08}(\textcolor{blue}{0.00})&26.04\\
			IU~\cite{1modelsparisty23jia}            &48.21\scalebox{0.7}{$\pm$0.52}(\textcolor{blue}{3.10})&94.95\scalebox{0.7}{$\pm$4.46}(\textcolor{blue}{94.95})&92.21\scalebox{0.7}{$\pm$0.21}(\textcolor{blue}{7.78})&20.98\scalebox{0.7}{$\pm$2.29}(\textcolor{blue}{19.31})&30.32\\
			BU~\cite{chen2023boundary} &50.20\scalebox{0.7}{$\pm$1.49}(\textcolor{blue}{1.11})&15.35\scalebox{0.7}{$\pm$1.23}(\textcolor{blue}{15.35})&88.73\scalebox{0.7}{$\pm$0.96}(\textcolor{blue}{11.26})&15.40\scalebox{0.7}{$\pm$0.45}(\textcolor{blue}{13.73})&29.92\\ \hline
			SCRUB~\cite{kurmanji2023towards} &51.83\scalebox{0.7}{$\pm$0.19}(\textcolor{blue}{0.52})&1.64\scalebox{0.7}{$\pm$0.14}(\textcolor{blue}{1.64})&97.54\scalebox{0.7}{$\pm$0.62}(\textcolor{blue}{2.45})&5.00\scalebox{0.7}{$\pm$0.72}(\textcolor{blue}{3.33})&131.58\\
			$l1$ Sparse\cite{1modelsparisty23jia} &51.34\scalebox{0.7}{$\pm$0.18}(\textcolor{blue}{0.03})&0.00\scalebox{0.7}{$\pm$0.00}(\textcolor{blue}{0.00})&99.93\scalebox{0.7}{$\pm$0.02}(\textcolor{blue}{0.06})&15.40\scalebox{0.7}{$\pm$0.62}(\textcolor{blue}{13.73})&142.95\\
			SalUn~\cite{fan2024salun} &51.05\scalebox{0.7}{$\pm$0.21}(\textcolor{blue}{0.28})&3.05\scalebox{0.7}{$\pm$0.15}(\textcolor{blue}{3.05})&99.96\scalebox{0.7}{$\pm$0.01}(\textcolor{blue}{0.03})&1.67\scalebox{0.7}{$\pm$0.52}(\textcolor{blue}{0.00})&67.41\\
			SFRon~\cite{huang2024unified} &48.18\scalebox{0.7}{$\pm$0.25}(\textcolor{blue}{3.13})&0.00\scalebox{0.7}{$\pm$0.00}(\textcolor{blue}{0.00})&95.63\scalebox{0.7}{$\pm$0.52}(\textcolor{blue}{4.36})&87.43\scalebox{0.7}{$\pm$2.35}(\textcolor{blue}{85.76})&95.69\\
			RUM~\cite{zhao2024what} &51.79\scalebox{0.7}{$\pm$0.12}(\textcolor{blue}{0.46})&2.85\scalebox{0.7}{$\pm$0.28}(\textcolor{blue}{2.85})&99.88\scalebox{0.7}{$\pm$0.02}(\textcolor{blue}{0.03})&0.00\scalebox{0.7}{$\pm$0.00}(\textcolor{blue}{1.67})&98.28\\ \hline
			OUR &51.98\scalebox{0.7}{$\pm$0.13}(\textcolor{blue}{0.67})&0.00\scalebox{0.7}{$\pm$0.00}(\textcolor{blue}{0.00})&99.97\scalebox{0.7}{$\pm$0.01}(\textcolor{blue}{0.02})&3.67\scalebox{0.7}{$\pm$0.11}(\textcolor{blue}{0.00})&78.24\\ \hline\hline
			\rowcolor{gray!7}
			\multicolumn{6}{c}{CIFAR100 (-5) - ResNet-50}\\ \hline
			Retrain    &62.63\scalebox{0.7}{$\pm$0.14}(\textcolor{blue}{0.00})&0.00\scalebox{0.7}{$\pm$0.00}(\textcolor{blue}{0.00})&99.98\scalebox{0.7}{$\pm$0.00}(\textcolor{blue}{0.00})&0.00\scalebox{0.7}{$\pm$0.00}(\textcolor{blue}{0.00})&2718.51\\
			FT~\cite{amnesiac21graves}           &63.90\scalebox{0.7}{$\pm$0.29}(\textcolor{blue}{1.27})&0.63\scalebox{0.7}{$\pm$0.31}(\textcolor{blue}{0.63})&99.98\scalebox{0.7}{$\pm$0.00}(\textcolor{blue}{0.00})&0.00\scalebox{0.7}{$\pm$0.00}(\textcolor{blue}{0.00})&123.46\\
			GA~\cite{thudi2022unrolling}    		&61.02\scalebox{0.7}{$\pm$0.19}(\textcolor{blue}{1.61})&0.00\scalebox{0.7}{$\pm$0.00}(\textcolor{blue}{0.00})&99.96\scalebox{0.7}{$\pm$0.02}(\textcolor{blue}{0.02})&14.15\scalebox{0.7}{$\pm$2.24}(\textcolor{blue}{14.15})&57.48\\
			RL~\cite{amnesiac21graves}           &63.84\scalebox{0.7}{$\pm$0.15}(\textcolor{blue}{1.21})&0.00\scalebox{0.7}{$\pm$0.08}(\textcolor{blue}{0.00})&99.98\scalebox{0.7}{$\pm$0.01}(\textcolor{blue}{0.00})&6.33\scalebox{0.7}{$\pm$0.68}(\textcolor{blue}{6.33})&42.58\\
			IU~\cite{1modelsparisty23jia}   		 &61.09\scalebox{0.7}{$\pm$0.23}(\textcolor{blue}{1.54})&0.44\scalebox{0.7}{$\pm$0.05}(\textcolor{blue}{0.44})&99.95\scalebox{0.7}{$\pm$0.02}(\textcolor{blue}{0.03})&0.00\scalebox{0.7}{$\pm$0.00}(\textcolor{blue}{0.00})&16.76\\
			BU~\cite{chen2023boundary} &61.15\scalebox{0.7}{$\pm$0.14}(\textcolor{blue}{1.48})&0.00\scalebox{0.7}{$\pm$0.00}(\textcolor{blue}{0.00})&89.55\scalebox{0.7}{$\pm$3.21}(\textcolor{blue}{10.43})&0.00\scalebox{0.7}{$\pm$0.00}(\textcolor{blue}{0.00})&49.15\\ \hline
			SCRUB~\cite{kurmanji2023towards} &61.02\scalebox{0.7}{$\pm$0.35}(\textcolor{blue}{1.61})&0.00\scalebox{0.7}{$\pm$0.00}(\textcolor{blue}{0.00})&99.97\scalebox{0.7}{$\pm$0.02}(\textcolor{blue}{0.01})&0.00\scalebox{0.7}{$\pm$0.00}(\textcolor{blue}{0.00})&136.81\\
			$l1$ Sparse\cite{1modelsparisty23jia} &62.13\scalebox{0.7}{$\pm$0.16}(\textcolor{blue}{0.50})&0.00\scalebox{0.7}{$\pm$0.00}(\textcolor{blue}{0.00})&98.47\scalebox{0.7}{$\pm$0.14}(\textcolor{blue}{1.52})&1.33\scalebox{0.7}{$\pm$0.11}(\textcolor{blue}{1.33})&126.67\\
			SalUn~\cite{fan2024salun} &62.93\scalebox{0.7}{$\pm$0.18}(\textcolor{blue}{0.30})&0.00\scalebox{0.7}{$\pm$0.00}(\textcolor{blue}{0.00})&98.90\scalebox{0.7}{$\pm$0.08}(\textcolor{blue}{0.10})&6.33\scalebox{0.7}{$\pm$2.43}(\textcolor{blue}{6.33})&38.29\\
			SFRon~\cite{huang2024unified} &61.35\scalebox{0.7}{$\pm$0.12}(\textcolor{blue}{0.28})&0.00\scalebox{0.7}{$\pm$0.00}(\textcolor{blue}{0.00})&95.76\scalebox{0.7}{$\pm$1.14}(\textcolor{blue}{0.04})&29.33\scalebox{0.7}{$\pm$5.47}(\textcolor{blue}{29.33})&150.77\\ 
			RUM~\cite{zhao2024what} &62.24\scalebox{0.7}{$\pm$0.25}(\textcolor{blue}{0.39})&0.00\scalebox{0.7}{$\pm$0.00}(\textcolor{blue}{0.00})&99.07\scalebox{0.7}{$\pm$0.12}(\textcolor{blue}{0.91})&0.00\scalebox{0.7}{$\pm$0.00}(\textcolor{blue}{0.00})&136.44\\ \hline
			OUR &62.88\scalebox{0.7}{$\pm$0.13}(\textcolor{blue}{0.25})&0.00\scalebox{0.7}{$\pm$0.00}(\textcolor{blue}{0.00})&99.96\scalebox{0.7}{$\pm$0.02}(\textcolor{blue}{0.02})&0.00\scalebox{0.7}{$\pm$0.00}(\textcolor{blue}{0.00})&72.12\\ \hline
			\multicolumn{6}{c}{\textbf{10\% Random Sample Unlearning (Sample-wise Unlearning)}}\\ \hline
			\makecell{MU}&\makecell{TA/\%}&\makecell{UA/\%}&\makecell{RA/\%}&\makecell{MIA Efficacy/\%} &\makecell{RTE (s)}\\ \hline
			\rowcolor{gray!7}
			\multicolumn{6}{c}{CIFAR10 - ResNet-18}\\ \hline
			Retrain &88.98\scalebox{0.7}{$\pm$0.39}(\textcolor{blue}{0.00})&88.91\scalebox{0.7}{$\pm$0.51}(\textcolor{blue}{0.00})&99.75\scalebox{0.7}{$\pm$0.21}(\textcolor{blue}{0.00})&6.40\scalebox{0.7}{$\pm$0.25}(\textcolor{blue}{0.00})&2401.89\\
			FT~\cite{amnesiac21graves}        &87.76\scalebox{0.7}{$\pm$0.26}(\textcolor{blue}{1.12})&88.36\scalebox{0.7}{$\pm$0.51}(\textcolor{blue}{0.55})&97.39\scalebox{0.7}{$\pm$0.25}(\textcolor{blue}{2.36})&6.80\scalebox{0.7}{$\pm$0.31}(\textcolor{blue}{0.40})&145.52\\
			GA~\cite{thudi2022unrolling}        &89.79\scalebox{0.7}{$\pm$0.26}(\textcolor{blue}{0.81})&85.73\scalebox{0.7}{$\pm$2.15}(\textcolor{blue}{0.55})&98.09\scalebox{0.7}{$\pm$0.06}(\textcolor{blue}{1.66})&11.00\scalebox{0.7}{$\pm$0.37}(\textcolor{blue}{4.60})&116.41\\
			RL~\cite{amnesiac21graves}         &88.78\scalebox{0.7}{$\pm$0.28}(\textcolor{blue}{0.10})&89.39\scalebox{0.7}{$\pm$0.17}(\textcolor{blue}{0.48})&96.38\scalebox{0.7}{$\pm$0.13}(\textcolor{blue}{4.37})&8.00\scalebox{0.7}{$\pm$0.42}(\textcolor{blue}{1.60})&43.65\\
			IU~\cite{1modelsparisty23jia}         &85.70\scalebox{0.7}{$\pm$0.21}(\textcolor{blue}{3.28})&90.02\scalebox{0.7}{$\pm$0.43}(\textcolor{blue}{1.11})&91.11\scalebox{0.7}{$\pm$0.79}(\textcolor{blue}{8.64})&12.20\scalebox{0.7}{$\pm$1.50}(\textcolor{blue}{5.80})&30.32\\
			FF~\cite{2eternal20golatkar}		 &83.99\scalebox{0.7}{$\pm$0.19}(\textcolor{blue}{4.99})&90.39\scalebox{0.7}{$\pm$0.25}(\textcolor{blue}{1.48})&90.80\scalebox{0.7}{$\pm$0.36}(\textcolor{blue}{8.85})&9.60\scalebox{0.7}{$\pm$0.14}(\textcolor{blue}{3.20})&2161.07\\ \hline
			SCRUB~\cite{kurmanji2023towards} &89.40\scalebox{0.7}{$\pm$0.15}(\textcolor{blue}{0.42})&89.73\scalebox{0.7}{$\pm$0.21}(\textcolor{blue}{0.83})&97.19\scalebox{0.7}{$\pm$0.04}(\textcolor{blue}{2.56})&8.40\scalebox{0.7}{$\pm$0.39}(\textcolor{blue}{2.00})&1353.33\\
			$l1$ Sparse\cite{1modelsparisty23jia}	&85.91\scalebox{0.7}{$\pm$0.21}(\textcolor{blue}{4.07})&87.03\scalebox{0.7}{$\pm$0.83}(\textcolor{blue}{1.88})&94.33\scalebox{0.7}{$\pm$0.14}(\textcolor{blue}{5.42})&6.20\scalebox{0.7}{$\pm$0.08}(\textcolor{blue}{0.20})&140.06\\
			SalUn~\cite{fan2024salun}	&89.89\scalebox{0.7}{$\pm$0.25}(\textcolor{blue}{0.91})&89.23\scalebox{0.7}{$\pm$0.21}(\textcolor{blue}{0.32})&97.71\scalebox{0.7}{$\pm$0.08}(\textcolor{blue}{2.04})&7.20\scalebox{0.7}{$\pm$0.13}(\textcolor{blue}{0.80})&32.77\\
			SFRon~\cite{huang2024unified}		&90.58\scalebox{0.7}{$\pm$0.28}(\textcolor{blue}{1.60})&86.69\scalebox{0.7}{$\pm$0.79}(\textcolor{blue}{2.22})&99.02\scalebox{0.7}{$\pm$0.01}(\textcolor{blue}{0.73})&13.20\scalebox{0.7}{$\pm$1.25}(\textcolor{blue}{6.80})&114.39\\ 
			RUM~\cite{zhao2024what}	&91.16\scalebox{0.7}{$\pm$0.27}(\textcolor{blue}{2.18})&88.04\scalebox{0.7}{$\pm$0.34}(\textcolor{blue}{0.87})&99.97\scalebox{0.7}{$\pm$0.01}(\textcolor{blue}{0.22})&4.20\scalebox{0.7}{$\pm$0.09}(\textcolor{blue}{2.00})&110.59\\ \hline
			OUR	&88.38\scalebox{0.7}{$\pm$0.19}(\textcolor{blue}{0.60})&88.99\scalebox{0.7}{$\pm$0.22}(\textcolor{blue}{0.08})&99.23\scalebox{0.7}{$\pm$0.14}(\textcolor{blue}{0.52})&7.00\scalebox{0.7}{$\pm$0.31}(\textcolor{blue}{0.60})&91.00\\ \hline
			\rowcolor{gray!7}
			\multicolumn{6}{c}{CIFAR20 - ViT}\\ \hline
			Retrain    &69.16\scalebox{0.7}{$\pm$0.24}(\textcolor{blue}{0.00})&69.07\scalebox{0.7}{$\pm$0.18}(\textcolor{blue}{0.00})&99.99\scalebox{0.7}{$\pm$0.00}(\textcolor{blue}{0.00})&12.20\scalebox{0.7}{$\pm$0.35}(\textcolor{blue}{0.00})&2313.38\\
			FT~\cite{amnesiac21graves}            &64.00\scalebox{0.7}{$\pm$0.21}(\textcolor{blue}{5.16})&68.50\scalebox{0.7}{$\pm$0.25}(\textcolor{blue}{0.57})&94.29\scalebox{0.7}{$\pm$0.34}(\textcolor{blue}{5.46})&13.80\scalebox{0.7}{$\pm$1.35}(\textcolor{blue}{1.60})&96.42\\
			GA~\cite{thudi2022unrolling}             &61.21\scalebox{0.7}{$\pm$0.15}(\textcolor{blue}{7.95})&74.77\scalebox{0.7}{$\pm$0.23}(\textcolor{blue}{5.70})&80.66\scalebox{0.7}{$\pm$2.72}(\textcolor{blue}{19.09})&24.80\scalebox{0.7}{$\pm$6.34}(\textcolor{blue}{12.60})&146\\
			RL~\cite{amnesiac21graves}      &63.69\scalebox{0.7}{$\pm$0.27}(\textcolor{blue}{5.47})&67.48\scalebox{0.7}{$\pm$0.13}(\textcolor{blue}{1.59})&88.45\scalebox{0.7}{$\pm$0.31}(\textcolor{blue}{11.30})&13.80\scalebox{0.7}{$\pm$1.25}(\textcolor{blue}{1.60})&71.29\\
			IU~\cite{1modelsparisty23jia}         &62.21\scalebox{0.7}{$\pm$0.15}(\textcolor{blue}{6.95})&82.24\scalebox{0.7}{$\pm$0.24}(\textcolor{blue}{13.17})&86.05\scalebox{0.7}{$\pm$0.76}(\textcolor{blue}{13.70})&31.40\scalebox{0.7}{$\pm$3.58}(\textcolor{blue}{19.20})&57.41\\
			FF~\cite{2eternal20golatkar} &60.19\scalebox{0.7}{$\pm$0.08}(\textcolor{blue}{8.97})&88.24\scalebox{0.7}{$\pm$0.13}(\textcolor{blue}{19.17})&88.10\scalebox{0.7}{$\pm$0.45}(\textcolor{blue}{11.70})&33.24\scalebox{0.7}{$\pm$4.53}(\textcolor{blue}{21.40})&1680.61\\ \hline
			SCRUB~\cite{kurmanji2023towards} &62.72\scalebox{0.7}{$\pm$0.21}(\textcolor{blue}{6.44})&73.51\scalebox{0.7}{$\pm$1.43}(\textcolor{blue}{4.44})&76.58\scalebox{0.7}{$\pm$1.65}(\textcolor{blue}{23.17})&17.60\scalebox{0.7}{$\pm$1.32}(\textcolor{blue}{5.40})&194.61\\
			$l1$ Sparse\cite{1modelsparisty23jia}	&62.93\scalebox{0.7}{$\pm$0.09}(\textcolor{blue}{6.23})&68.01\scalebox{0.7}{$\pm$0.11}(\textcolor{blue}{1.06})&90.49\scalebox{0.7}{$\pm$0.23}(\textcolor{blue}{9.26})&14.00\scalebox{0.7}{$\pm$0.85}(\textcolor{blue}{1.80})&83.30\\
			SalUn~\cite{fan2024salun} 	&63.45\scalebox{0.7}{$\pm$0.27}(\textcolor{blue}{5.71})&65.79\scalebox{0.7}{$\pm$0.31}(\textcolor{blue}{3.28})&92.55\scalebox{0.7}{$\pm$0.18}(\textcolor{blue}{7.20})&12.00\scalebox{0.7}{$\pm$1.75}(\textcolor{blue}{0.20})&133.19\\
			SFRon~\cite{huang2024unified}		&66.20\scalebox{0.7}{$\pm$0.24}(\textcolor{blue}{2.96})&69.00\scalebox{0.7}{$\pm$0.45}(\textcolor{blue}{0.07})&97.60\scalebox{0.7}{$\pm$0.08}(\textcolor{blue}{2.15})&21.80\scalebox{0.7}{$\pm$3.25}(\textcolor{blue}{9.60})&53.85\\
			RUM~\cite{zhao2024what}	&64.87\scalebox{0.7}{$\pm$0.18}(\textcolor{blue}{4.29})&67.13\scalebox{0.7}{$\pm$0.24}(\textcolor{blue}{1.94})&91.36\scalebox{0.7}{$\pm$0.43}(\textcolor{blue}{8.39})&11.00\scalebox{0.7}{$\pm$0.76}(\textcolor{blue}{1.20})&103.10\\ \hline
			OUR 	&67.90\scalebox{0.7}{$\pm$0.21}(\textcolor{blue}{1.92})&69.37\scalebox{0.7}{$\pm$0.19}(\textcolor{blue}{0.30})&97.72\scalebox{0.7}{$\pm$0.08}(\textcolor{blue}{0.44})&12.80\scalebox{0.7}{$\pm$1.24}(\textcolor{blue}{0.60})&187.56\\ \hline
		\end{tabular}
	\end{spacing}
\vspace{-0.155cm}
\end{table}

\begin{table*}
	\centering
	\caption{Performance of Attacks on Class-wise Unlearning: Inferring the Membership of Unlearned Classes from OOD Classes}
	\vspace{-0.28cm}
	\label{tb:overall_evaluation_class}
	\tiny
	\begin{spacing}{1.0}
		\begin{tabular}{|p{0.88cm}<{\centering}|p{0.88cm}<{\centering}p{0.96cm}<{\centering}p{0.88cm}<{\centering}p{0.95cm}<{\centering}p{0.95cm}<{\centering}p{0.95cm}<{\centering}|p{0.88cm}<{\centering}p{0.88cm}<{\centering}p{0.88cm}<{\centering}p{0.88cm}<{\centering}p{0.95cm}<{\centering}|p{0.88cm}<{\centering}|}
			\hline
			\multirow{2}{*}{\makecell{Attacks}}& \multicolumn{12}{c|}{Balanced Accuracy (TPR @0.1 FPR) / \%}\\ \cline{2-13}
			&\makecell{Retrain}&\makecell{FT}&\makecell{GA}&\makecell{RL}&\makecell{IU}&\makecell{BU}&\makecell{SCRUB}&\makecell{$l1$ Sparse}&\makecell{SalUn}&\makecell{SFRon}&\makecell{RUM}&\makecell{OUR}\\ \hline
			\rowcolor{gray!7}
			\multicolumn{13}{|c|}{\textbf{CIFAR20 (-5)}}\\ \hline
			MIA-LiRA&52.28(31.25)&59.04(18.75)&61.57(37.50)&50.00(31.25)&64.70(18.75)&52.53(25.00)&50.42 (6.25)&50.00(25.00)&59.29(6.25)&\textbf{89.02(87.50})&61.49(25.00)&\textbf{52.45}(6.25)
			\\
			MIA-UP&50.00(18.75)&56.84(12.50)&64.70(37.50)&61.40(18.75)&69.00(50.00)&64.95(12.50)&50.00(0.00)&50.25(18.75)&67.40(50.00)&73.14(43.75)&50.00(18.75)&50.00(18.75)
			\\
			ReA&50.00(6.25)&\textbf{86.99(75.00)}&\textbf{71.03(43.75)}&\textbf{95.52(93.75)}&\textbf{89.02(87.50)}&\textbf{91.05(93.75)}&\textbf{68.75(31.25)}&\textbf{63.34(25.00)}&\textbf{95.27(100.00)}&84.29(81.25)&\textbf{85.64(81.25)}&52.28(\textbf{12.50})\\ \hline
			\rowcolor{gray!7}
			\multicolumn{13}{|c|}{\textbf{CIFAR100 (-5)}}\\
			\hline
			MIA-LiRA&50.00(6.25)&57.52(6.25)&55.49(25.00)&69.00(25.00)&87.67(87.50)&81.42(75.00)&\textbf{64.44(31.25)}&59.54(18.75)&69.17(43.75)&65.37(37.50)&66.89(37.50)&50.00(6.50)
			\\
			MIA-UP&50.08(12.50)&63.60(43.75)&59.54(18.75)&56.59(25.00)&63.09 (25.00)&67.23(37.50)&52.70(6.25)&55.49(25.00)&64.27(37.50)&\textbf{73.14(56.25)}&54.05(6.25)&\textbf{54.73(12.50)}\\
			ReA&50.00(6.25))&\textbf{86.32(75.00)}&\textbf{60.22(31.25)}&\textbf{82.47(81.50)}&\textbf{89.24(89.75)}&\textbf{84.17(83.75)}&61.57(31.25)&\textbf{62.67(25.00)}&\textbf{81.33(62.50)}&65.88(31.25)&\textbf{82.77(87.50)}&50.68(0.00)
			\\ \hline
		\end{tabular}
	\end{spacing}
	\vspace{-0.15cm}
\end{table*}

\begin{table*}
	\centering
	\caption{Performance of Attacks on Random Sample Forgetting (10\%): Infer the Membership of Unlearned Samples}
	\vspace{-0.28cm}
	\label{tb:overall_evaluation_sample}
	\tiny
	\begin{spacing}{1.0}
		%\begin{tabular}{|p{1.22cm}<{\centering}|p{1.7cm}<{\centering}|p{1.15cm}<{\centering}|p{0.86cm}<{\centering}|p{0.86cm}<{\centering}|p{0.86cm}<{\centering}|p{0.86cm}<{\centering}|p{1.15cm}<{\centering}|p{0.86cm}<{\centering}|p{0.86cm}<{\centering}|p{0.86cm}<{\centering}|p{0.86cm}<{\centering}|}
		%{|c|c|p{1cm}<{\centering}|p{1cm}<{\centering}|p{0.95cm}<{\centering}|p{1cm}<{\centering}|}
		\begin{tabular}{|p{0.88cm}<{\centering}|p{0.88cm}<{\centering}p{0.96cm}<{\centering}p{0.88cm}<{\centering}p{0.95cm}<{\centering}p{0.95cm}<{\centering}p{0.95cm}<{\centering}|p{0.88cm}<{\centering}p{0.88cm}<{\centering}p{0.88cm}<{\centering}p{0.88cm}<{\centering}p{0.95cm}<{\centering}|p{0.88cm}<{\centering}|}
			\hline
			\multirow{2}{*}{Attacks}& \multicolumn{12}{c|}{Balanced Accuracy (TPR@0.1FPR) / \%}\\ \cline{2-13}
			&Retrain&FT&GA&RL&IU&FF&SCRUB&$l1$ Sparse&SalUn\cite{fan2024salun}&SFRon&RUM&OUR\\ \hline
			\rowcolor{gray!7}
			\multicolumn{13}{|c|}{\textbf{CIFAR10}}\\ \hline
			MIA-LiRA&50.63(6.40)&50.60(6.80)&51.50(11.00)&51.50(8.00)&53.60(12.20)&53.60(9.60)&50.90(8.40)&50.90(6.20)&52.40(7.20)&52.20(13.20)&51.00(4.20)&50.69(7.00)\\
			MIA-UP&49.99(9.80)&\textbf{50.66(9.92)}&52.72(15.40)&50.00 (\textbf{10.24})&52.84(12.28)&52.50(12.12)&50.34(\textbf{9.72})&50.78(\textbf{9.96})&51.34(\textbf{11.44})&52.68(\textbf{14.36})&\textbf{52.82(15.56)}&\textbf{50.70(9.88)}\\
			ReA&50.30(7.00)&50.40(8.60)&\textbf{55.60(15.60)}&\textbf{52.20}(9.60)&\textbf{58.40(24.20)}&\textbf{57.90(15.00)}&\textbf{51.60}(6.00)&\textbf{51.20}(6.00)&\textbf{53.40}(6.80)&\textbf{52.80}(13.60)&51.20(8.20)&50.48(7.20)\\ \hline
			\rowcolor{gray!7}
			\multicolumn{13}{|c|}{\textbf{CIFAR20}}\\ \hline
			MIA-LiRA&51.20(12.20)&56.10(13.80)&61.80(24.80)&53.20(\textbf{13.80})&64.80(31.40)&68.20 (33.24)&57.60(17.60)&53.10(\textbf{14.00})&53.80(\textbf{12.00})&56.80(21.80)&52.10(\textbf{11.00})&53.20(\textbf{12.80})\\
			MIA-UP&51.39(9.24)&54.46(10.80)&57.62(12.88)&53.48(8.88)&60.14(12.92)&59.89(12.24)&56.04(11.52)&52.84(9.72)&52.02(8.00)&51.36(20.84)&51.36(8.96)&52.46(9.88)\\
			ReA&51.35(10.80)&\textbf{57.00}(\textbf{12.80})&\textbf{72.90(44.00)}&\textbf{57.10}(12.20)&\textbf{72.70(44.40)}&\textbf{75.30(45.20)}&\textbf{63.30(33.40)}&\textbf{57.60}(12.00)&\textbf{55.40}(9.80)&\textbf{61.70(28.00)}&\textbf{53.20}(10.00)&\textbf{53.80}(11.50)\\ \hline
		\end{tabular}
	\end{spacing}
\vspace{-0.17cm}
\end{table*}

%\begin{figure}
%	\centering
%	\subfloat[Retrain]{
%		\includegraphics[width=0.3\columnwidth]{figure/auc/sample_wise/retrain_auc.pdf}}
%	\subfloat[SCRUB~\cite{kurmanji2023towards}]{
%		\includegraphics[width=0.3\columnwidth]{figure/auc/sample_wise/scrub_auc.pdf}}
%	\subfloat[OUR]{
%		\includegraphics[width=0.3\columnwidth]{figure/auc/sample_wise/orthogonality_auc.pdf}}
%	\vspace{-0.25cm}
%	\caption{ROC Analysis of Sample-wise MIA (Selected Methods). Appendix~\ref{app:roc_full} provides ROC analysis for all AMU methods.}
%	\vspace{-0.15cm}
%	\label{fig:auc_sample_wise}
%\end{figure}

\subsection{Quantified Analysis of Unlearning Residuals}
\label{subsec:visual_reps}
We quantify residuals in MU benchmarks to explain why ReA succeeds, and assess OUR’s effectiveness in scrubbing them. To validate this, we use distinct analyses for class-wise and sample-wise residuals. For class-wise cases, we visualize representation spaces to detect class-specific residuals. For sample-wise cases, we quantify residuals by measuring the recovered unlearned accuracy (UA) by ReA.

\begin{figure}
	\vspace{-0.1cm}
	\centering
	\subfloat[Retrain]{
		\includegraphics[width=0.28\columnwidth]{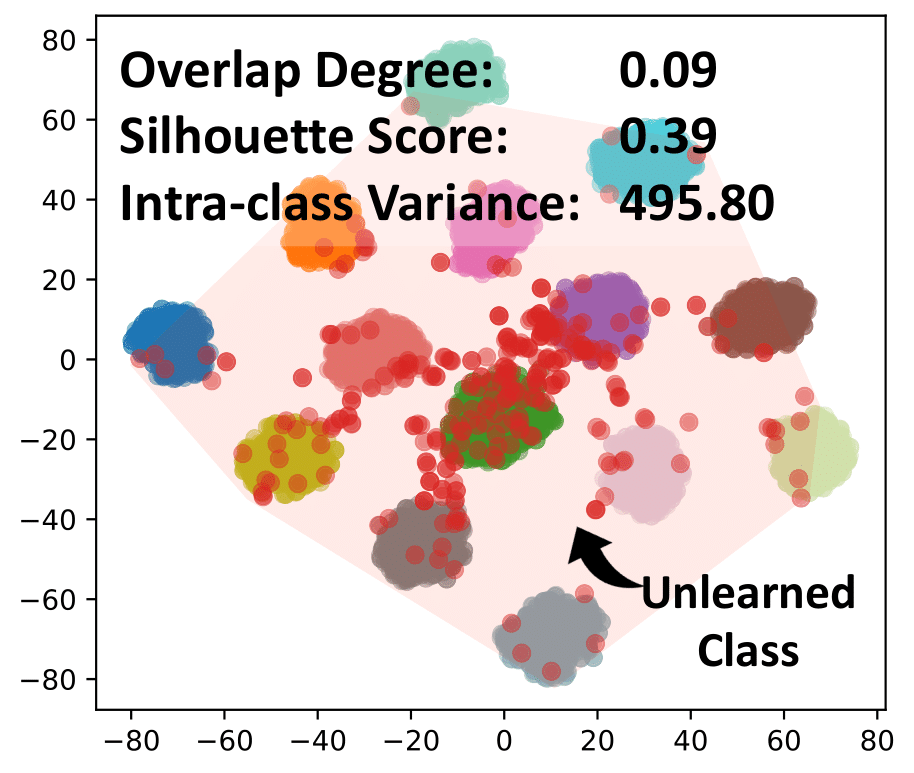}}~
	\subfloat[RUM~\cite{zhao2024what}]{
		\includegraphics[width=0.28\columnwidth]{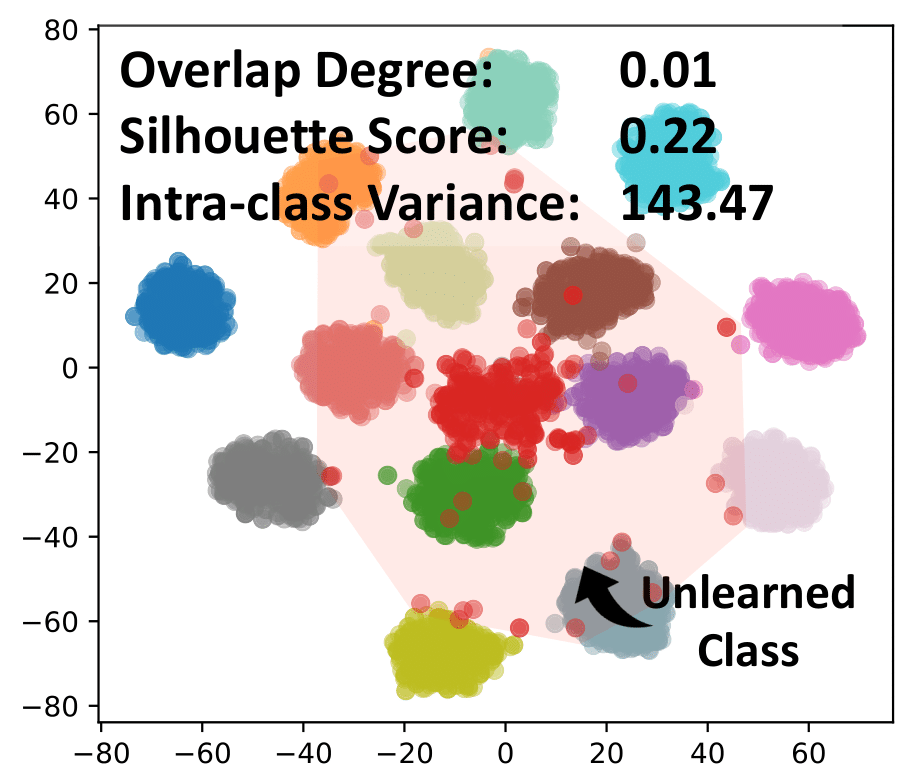}}~
	\subfloat[OUR]{
		\includegraphics[width=0.28\columnwidth]{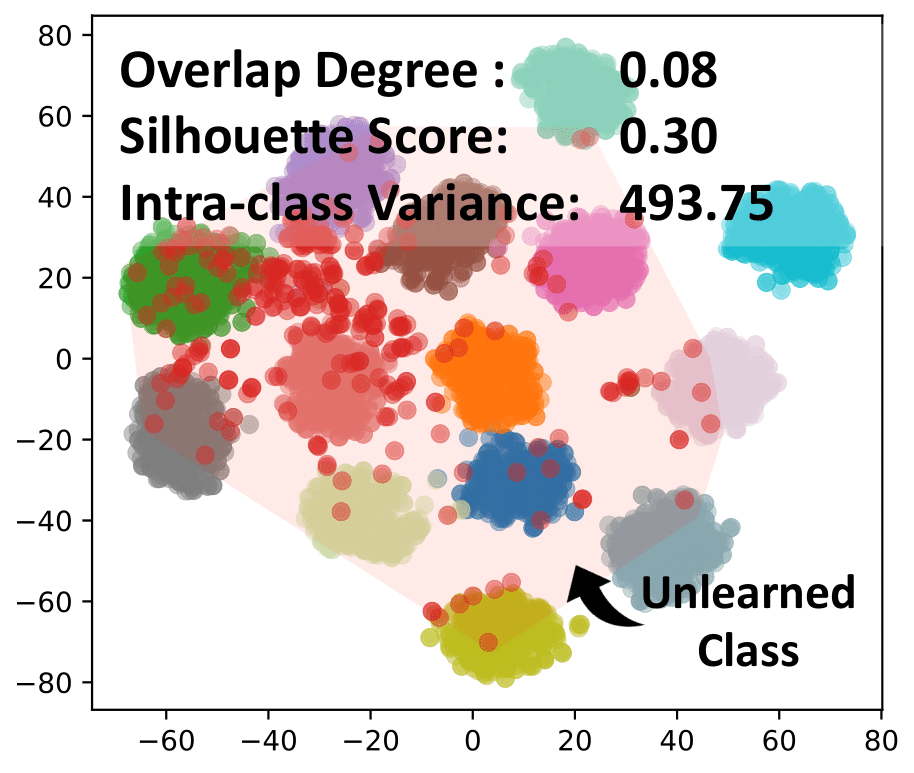}}
	% \vspace{-0.13cm}
	\vspace{-0.35cm}
	\caption{Representation space visualization of the unlearned model. OUR's treatment for the \textit{fruit \& vegetables} forget class (red) matches retrained models better than RUM. Appendix~\ref{app:representation_visual} provides results for all AMU methods.}
	\label{fig:tsne_ex}
		\vspace{-0.348cm}
\end{figure}

\begin{figure}
	\centering
	\subfloat[Class-wise Unlearning]{
		\includegraphics[width=0.43\columnwidth]{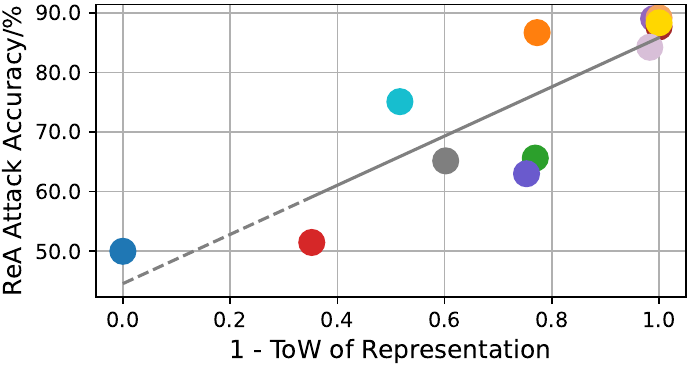}
		\label{fig:class-wise_analysis}}~~~
	\subfloat[Sample-wise Unlearning]{
		\includegraphics[width=0.43\columnwidth]{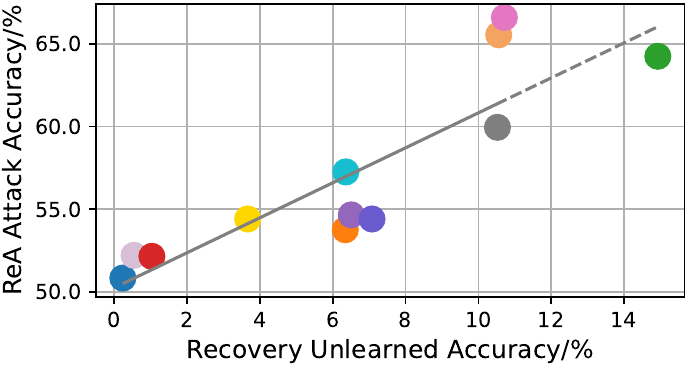}
		\label{fig:sample-wise_analysis}}\\
	\vspace{-0.05cm}
	\hspace{0.2cm}	\includegraphics[width=0.88\columnwidth]{tow_privacy_attack_legend.pdf}
	\vspace{-0.15cm}
	\caption{Quantified Residuals versus ReA Attack Accuracy.}
	\label{fig:relationship_rea}
		\vspace{-0.37cm}
\end{figure}

\noindent\textbf{Class-wise Residual Visualization}. 
Figure~\ref{fig:tsne_ex} compares t-SNE~\cite{van2008visualizing} projections of unlearned models in CIFAR-20 (-5) experiments and quantifies them with representation metrics: (1) \textbf{Overlap degree} (\(o\)) for class separation, (2) \textbf{Silhouette Score} (\(s\)) for cluster compactness, and (3) \textbf{Intra-class Variance} (\(\mathrm{var}\)). OUR achieves near-retrained model alignment ($\Delta\mathrm{var} <3$ vs. $352$ for RUM~\cite{zhao2024what}), indicating effective residual scrubbing. Figure~\ref{fig:class-wise_analysis} further reveals a strong correlation (Pearson $=0.88$) between representation misalignment (1-$\mathrm{ToW}(o, s, \mathrm{var})$) and ReA accuracy, confirming that representation residuals enable privacy breaches. Full metric definitions and results are in Appendix~\ref{app:representation_visual}.

\noindent\textbf{Sample-wise Residual Quantification}. 
For sample-wise cases, we assess residuals by tracking unlearning accuracy recovery during ReA. Figure~\ref{fig:sample-wise_analysis} shows that higher recovered accuracy ($\Delta\mathrm{UA}$) correlates with increased privacy leakage (Pearson $=0.87$). OUR minimizes both recovery ($\Delta\mathrm{UA} = 1.26\%$) and privacy risk (ReA$\leq 53.80\%$), while some methods like GA~\cite{thudi2022unrolling} exhibit strong recovery ($\Delta \text{UA} = 14.85\%$) and high vulnerability (ReA$\leq 72.90\%$). 

\subsection{Analysis of OUR Variants}
\label{sec:ex_variants}
To validate the necessity of OUR's core components, we evaluate four ablated variants: (1) \textit{Orth w/o Replay} (jointly optimizing orthogonal unlearning and utility preservation), (2) \textit{OUR w/o Sparsity} (removing $l1$ regularization during replay), (3) \textit{RL w/ Replay} (replacing orthogonal unlearning with random label (RL)~\cite{amnesiac21graves}), and (4) \textit{FT w/ Sparsity} (fine-tuning only with $l1$ regularization~\cite{1modelsparisty23jia}).

\begin{figure}
	\centering
	\subfloat[Class-wise Unlearning]{
		\includegraphics[width=0.46\columnwidth]{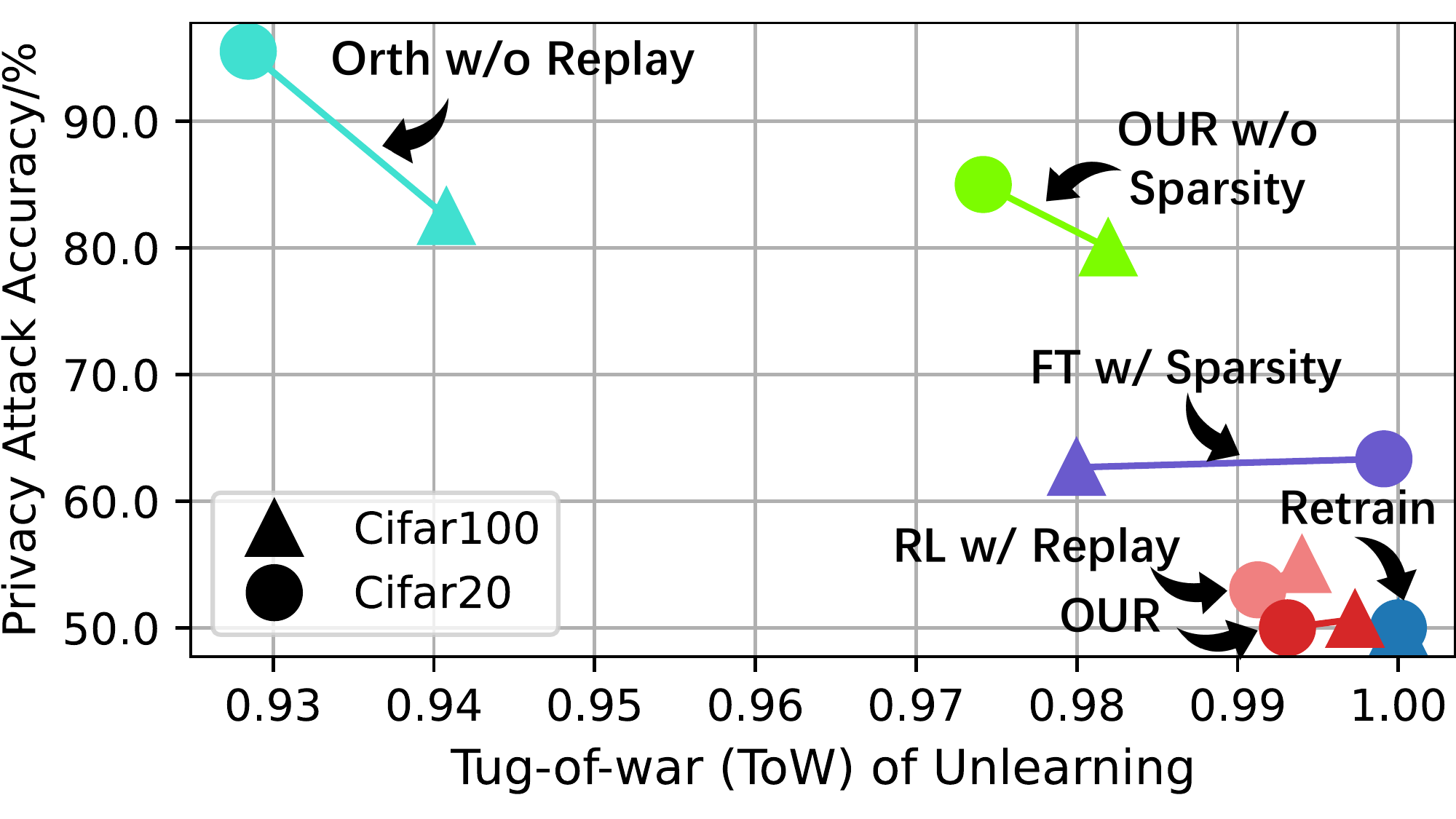}}~
	\subfloat[Sample-wise Unlearning]{
		\includegraphics[width=0.46\columnwidth]{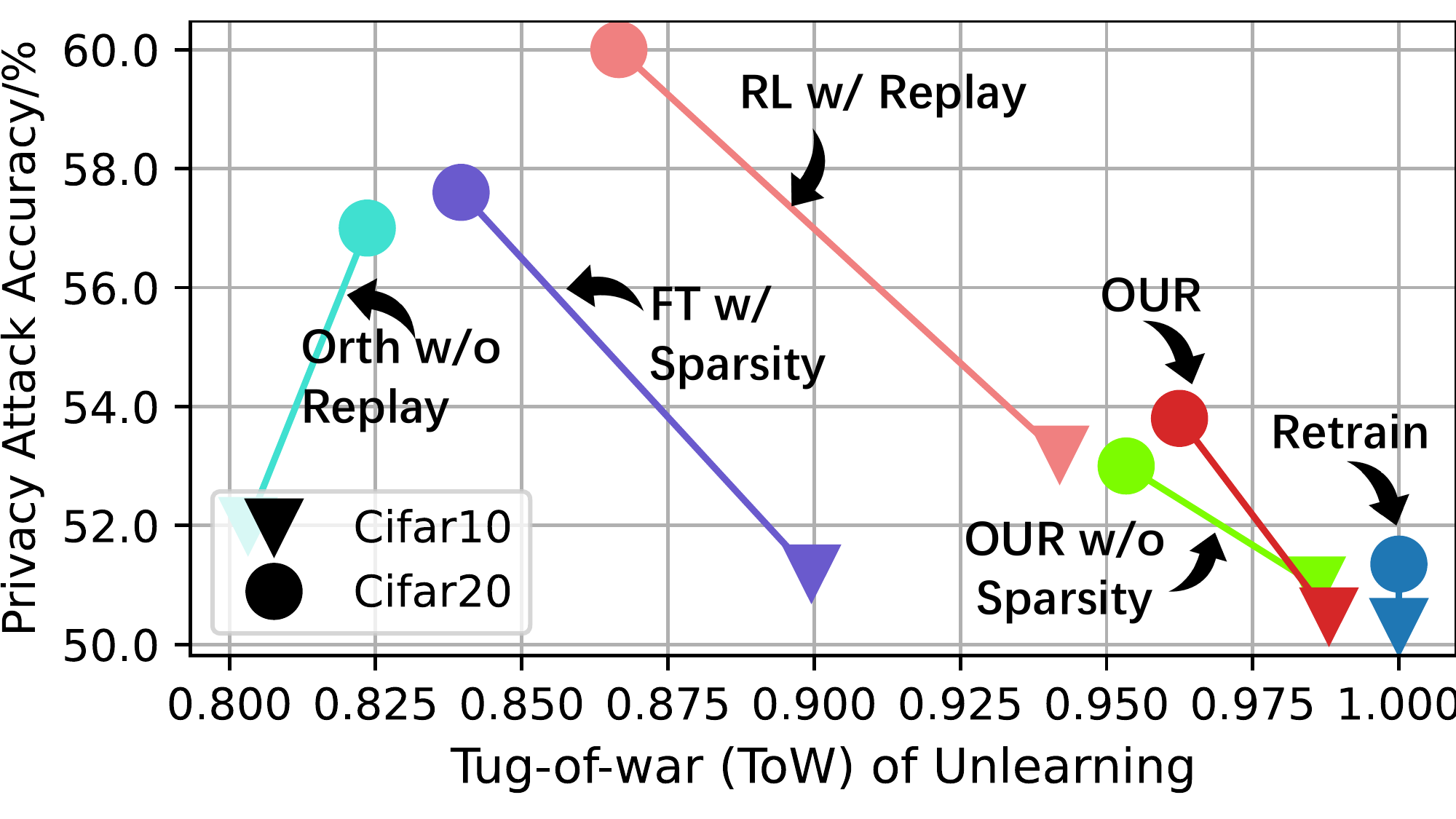}}
	\vspace{-0.2cm}
	\caption{Unlearning performance of OUR variants versus their privacy vulnerability test by ReA.}
	\vspace{-0.1cm}
	\label{fig:variants_tow_attack}
\end{figure}

Figure~\ref{fig:variants_tow_attack} plots unlearning efficacy (measured by ToW(UA,TA,RA)) against ReA privacy attack accuracy across datasets, and three critical observations emerge. First, the two-phase design is essential. \textit{Orth w/o Replay} performs the worst, which degrades unlearning efficacy by $0.06$ and increases ReA attack accuracy to $95.52\%$, showing that joint optimization in unlearning weakens both objectives. Second, orthogonal unlearning proves superior to alternatives. OUR reduces ReA accuracy by $2.89-6.12\%$ compared to \textit{RL w/ Replay}, as it removes more internal residuals. Third, $l1$ regularization in OUR enhances class-wise unlearning (reducing ReA accuracy by $35.07\%$), yet has a limited effect on sample-wise cases where the separated replay phase dominates. The synergy of orthogonal unlearning, two-phase design and $l1$ regularization achieves both effective unlearning and privacy protection.

\subsection{Experiments on Image Generation}  
\label{subsec:ex_gen}  
We evaluate privacy risks in identity-sensitive face generation using SD models. The target model synthesizes two identities (e.g., ``a photo of Laura'') with distinct features. Concept unlearning is performed to erase a target identity. Its unlearning efficacy is measured by Unlearning Identity Score Matching (ISM)~\cite{van2023anti}, abbreviated as \textbf{UISM} Retrained generation quality is assessed using three metrics: Retaining ISM (\textbf{RISM}) for identity accuracy, \textbf{FID}~\cite{fan2024salun} for distribution fidelity, and \textbf{BRISQUE}~\cite{madry2017towards} for artifacts. Appendix~\ref{subsec:app_gen} details setups and metrics and evaluates more unlearning benchmarks, e.g., SFRon~\cite{huang2024unified}. 

As shown in Table~\ref{tb:generation_mu}, OUR achieves near-retrained UISM with $12\%$ of retraining time but initially causes RISM degradation by $0.08$. To mitigate this, we integrate SalUn's parameter freezing strategy to preserve retained identity features, improving RISM by $0.06$ without compromising running-time efficiency. Visual results in Table~\ref{tb:image_mu} confirm that (SalUn-)OUR effectively removes target identities while maintaining retained output quality.

We design two MIA to evaluate MU's privacy leakage and report the results in Table~\ref{tb:rea_generation}: (1) \textbf{ReA}, adapted from class-wise ReA, and (2) \textbf{DiffAtk}~\cite{zhang2024generate}, which employs adversarial prompts to revive forgotten identities within limited iterations. To be practical, attackers are assumed to have access to public photos of the forgotten identity (not original training data) and lack knowledge of the forgotten prompt. Both attacks run for $100$ iterations, with success measured by the resonance index (Idx\textsubscript{r}) where ISM exceeds $0.70$. ReA proves most effective, requiring $25$ fewer Idx\textsubscript{r} in SalUn, while DiffAtk consistently fails. Second, orthogonal unlearning in OUR enhanced security for portrait generation compared to existing unlearning methods. OUR achieves a resilience index (Idx\textsubscript{r}) of $70$ (near-retrained models), even better than Meta-unlearning (Meta-Un)~\cite{gao2024meta}, which is designed to prevent forgotten concept recovery.

\begin{table}
	\caption{Performance of Identity-specific Concept Unlearning}
	\vspace{-0.18cm}
	\label{tb:generation_mu}
	\tiny
	\begin{spacing}{1.0}
		\begin{tabular}{p{0.8cm}<{\centering}|p{0.95cm}<{\centering}p{0.95cm}<{\centering}p{1.05cm}<{\centering}p{0.96cm}<{\centering}p{0.95cm}<{\centering}}
			\hline
			Metrics&Retrain&SalUn~\cite{fan2024salun}&Meta-Un~\cite{gao2024meta}&OUR&SalUn-OUR\\ \hline
			UISM&0.13\scalebox{0.7}{$\pm$0.02}(\textcolor{blue}{0.00})&0.49\scalebox{0.7}{$\pm$0.03}(\textcolor{blue}{0.36})&0.37\scalebox{0.7}{$\pm$0.05}(\textcolor{blue}{0.24})&0.16\scalebox{0.7}{$\pm$0.02}(\textcolor{blue}{0.03})&0.26\scalebox{0.7}{$\pm$0.01}(\textcolor{blue}{0.13})\\
			RISM&0.76\scalebox{0.7}{$\pm$0.02}(\textcolor{blue}{0.00})&0.71\scalebox{0.7}{$\pm$0.02}(\textcolor{blue}{0.05})&0.74\scalebox{0.7}{$\pm$0.04}(\textcolor{blue}{0.02})&0.68\scalebox{0.7}{$\pm$0.01}(\textcolor{blue}{0.08})&0.74\scalebox{0.7}{$\pm$0.02}(\textcolor{blue}{0.02})\\
			FID(\scalebox{0.8}{$\downarrow$})&101.82\scalebox{0.7}{$\pm$8.33}&108.23\scalebox{0.7}{$\pm$7.25}&122.76\scalebox{0.7}{$\pm$11.35}&110.31\scalebox{0.7}{$\pm$7.69}&103.55\scalebox{0.7}{$\pm$11.35}\\
			BRISQUE(\scalebox{0.8}{$\downarrow$})&10.76\scalebox{0.7}{$\pm$1.38}&15.15\scalebox{0.7}{$\pm$0.41}&15.02\scalebox{0.7}{$\pm$2.03}&15.58\scalebox{0.7}{$\pm$3.90}&12.03\scalebox{0.7}{$\pm$3.82}\\
			RTE (s)&937.50&239.31&485.31&110.25&113.83\\  \hline
		\end{tabular}	
		\hspace{-5cm}
	\end{spacing}	
\vspace{-0.02cm}
\end{table}

\begin{table}
	\caption{Images of Identity-specific Concept Unlearning}
	\vspace{-0.2cm}
	\label{tb:image_mu}
	\tiny
	\begin{tabular}{p{0.94cm}<\centering p{0.945cm}<\centering p{0.945cm}<\centering p{0.965cm}<\centering p{0.945cm}<\centering p{0.97cm}<\centering}
		\hline
		Target&Retrain&SalUn~\cite{fan2024salun}&Meta-Un~\cite{gao2024meta}&OUR&SalUn-OUR\\
		\hline
		\multicolumn{6}{c}{Unlearned Identity. Prompt: ``{{\fontfamily{pzc}\selectfont a photo of Sarah}}"}\\ 
		\includegraphics[width=1.21cm]{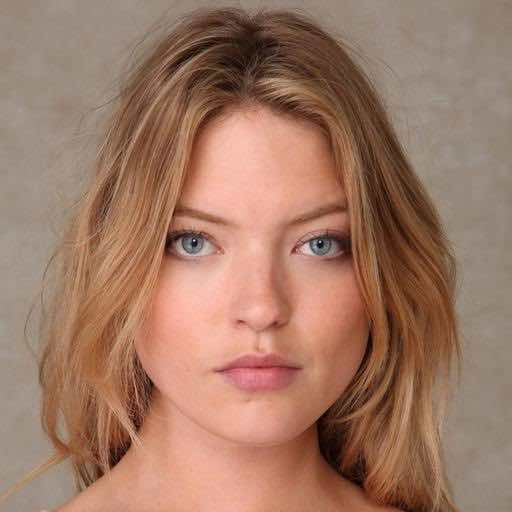}&\includegraphics[width=1.21cm]{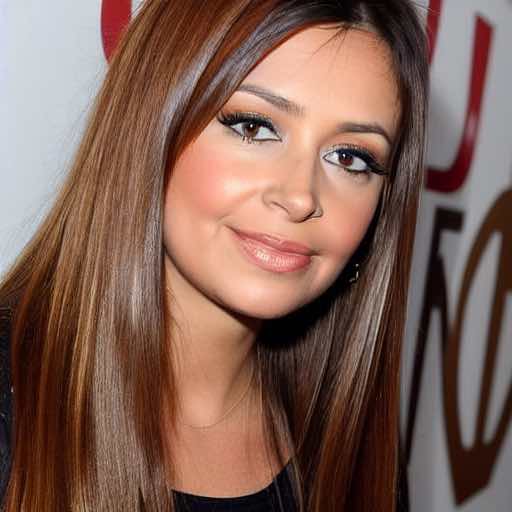}&\includegraphics[width=1.21cm]{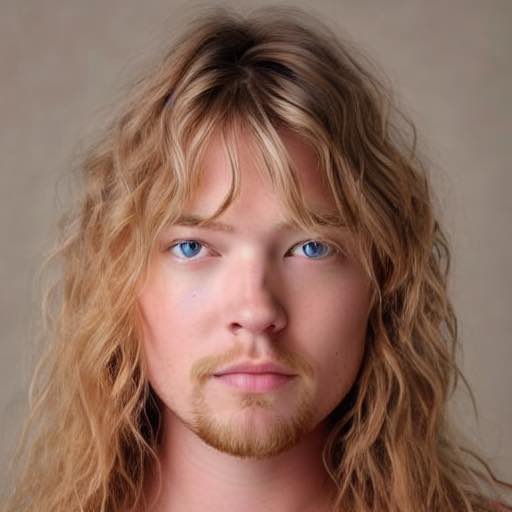}&\includegraphics[width=1.21cm]{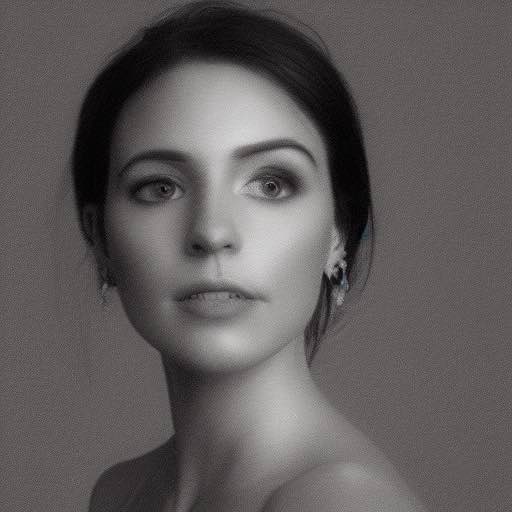}&\includegraphics[width=1.21cm]{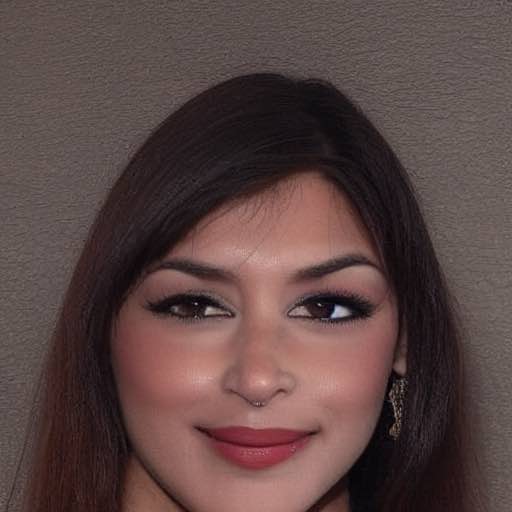}&\includegraphics[width=1.21cm]{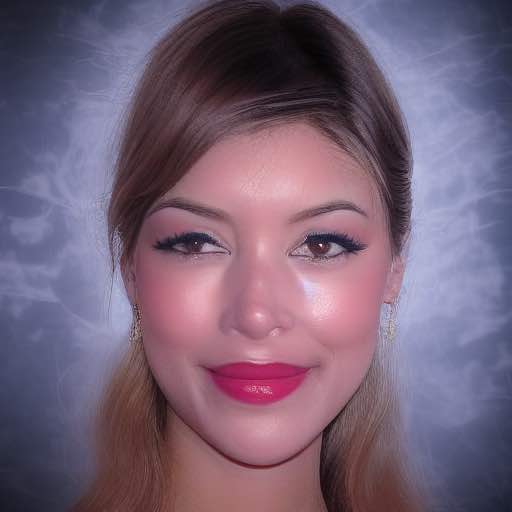}\\ \hline
		\multicolumn{6}{c}{Retained Identity. Prompt: ``{\fontfamily{pzc}\selectfont \textbf{a photo of Laura}}"}\\ 
		\includegraphics[width=1.21cm]{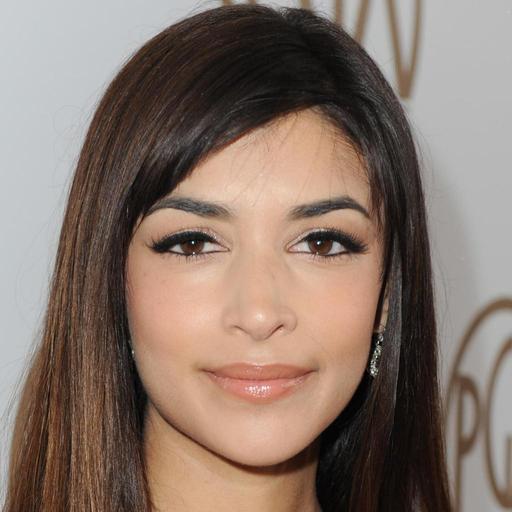}&\includegraphics[width=1.21cm]{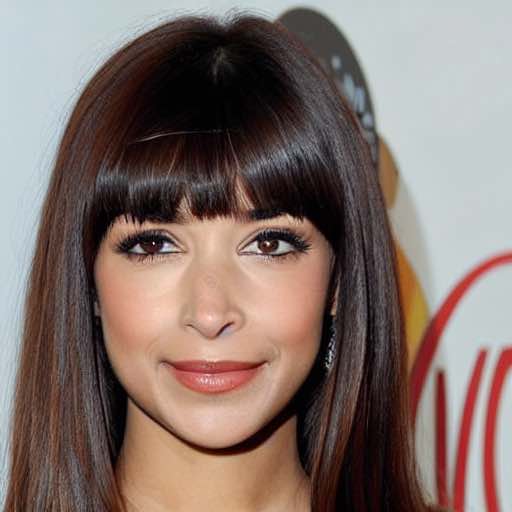}&\includegraphics[width=1.21cm]{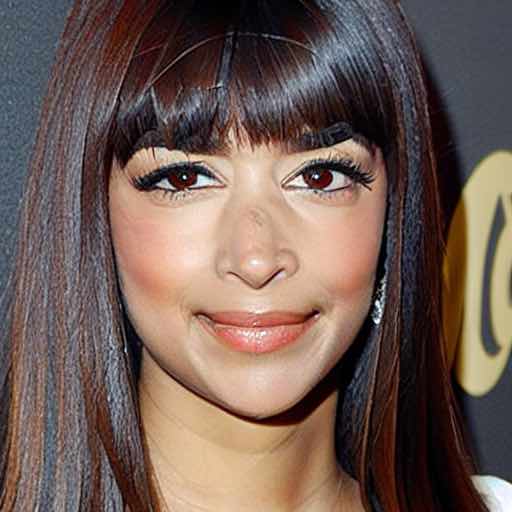}&\includegraphics[width=1.21cm]{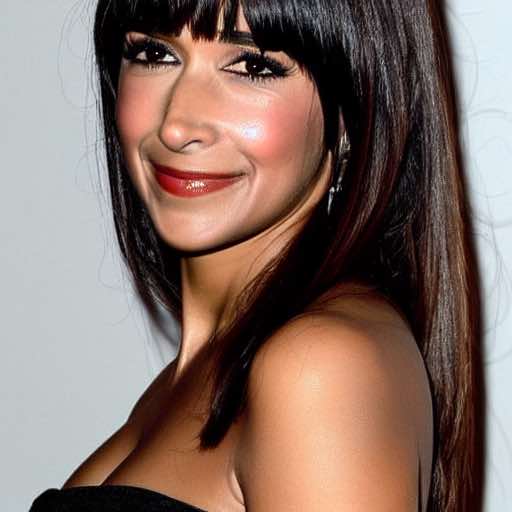}&\includegraphics[width=1.21cm]{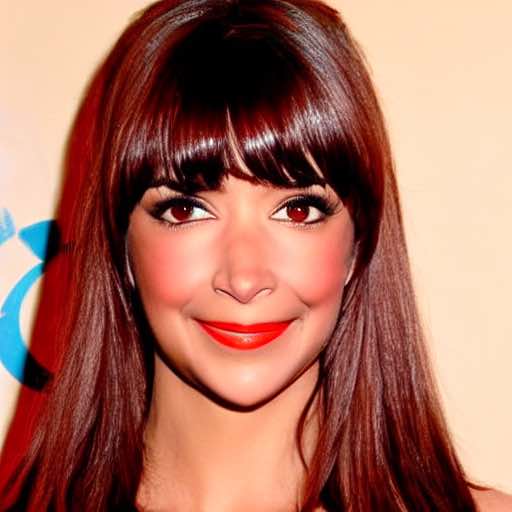}&\includegraphics[width=1.21cm]{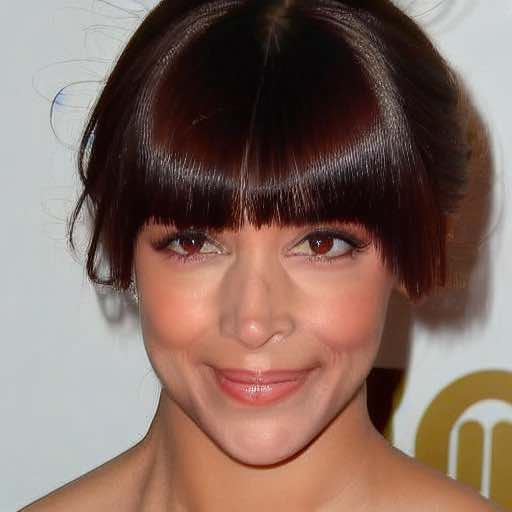}\\ \hline
	\end{tabular}
\end{table}

\begin{table}
	\caption{Privacy Attack against Concept Unlearning}
	\vspace{-0.22cm}
	\label{tb:rea_generation}
	\tiny
	\begin{spacing}{1.0}
		\begin{tabular}{|p{0.75cm}<{\centering}|p{0.94cm}<{\centering}p{0.94cm}<{\centering}p{1.05cm}<{\centering}p{1.05cm}<{\centering}p{0.95cm}<{\centering}|}
			\hline
			\multirow{2}{*}{Attacks}&\multicolumn{5}{c|}{Resonance Index, Idx\textsubscript{r}, (Maximum = 10) \scalebox{0.85}{$\times$ 10} @ ISM} \\ \cline{2-6}
			&Retrain&SalUn~\cite{fan2024salun}&Meta-Un~\cite{gao2024meta}&OUR&SalUn-OUR\\ \hline
			DiffAtk\cite{zhang2024generate}&10\scalebox{0.7}{$\pm$0.0}@0.18&10\scalebox{0.7}{$\pm$0.0}@0.10&10\scalebox{0.7}{$\pm$0.0}@0.09&10\scalebox{0.7}{$\pm$0.0}@0.10&10\scalebox{0.7}{$\pm$0.0}@0.08\\ \hline
			ReA&6.5\scalebox{0.7}{$\pm$1.0}@0.71&2\scalebox{0.7}{$\pm$0.5}@0.77&4.5\scalebox{0.7}{$\pm$1}@0.71&7.5\scalebox{0.7}{$\pm$1.0}@0.70&7\scalebox{0.7}{$\pm$1}@0.72\\ \hline
		\end{tabular}
	\end{spacing}
\vspace{-0.3cm}
\end{table}

\vspace{-0.2cm}
\section{Related Works}
\vspace{-0.1cm}
\label{sec:related_works}
\noindent\textbf{Privacy Hazards in Machine Unlearning}.
The growing adoption of machine unlearning (MU) has raised concerns about its unintended privacy risks. Current research identifies two primary leakage sources: (1) updates between model versions before and after unlearning~\cite{when21chen,hu2024learn,bertran2025reconstruction}, and (2) knowledge residuals caused by approximate unlearning algorithms~\cite{kurmanji2023towards}. While version control addresses the first, the second reveals more profound flaws: imperfect alignment between unlearned and test data creates measurable discrepancies. Crucially, we reveal that these discrepancies are not merely superficial; they manifest as residuals in the loss landscape. By exposing these hidden residuals, our work shows that existing unlearning methods inadvertently leak privacy. This highlights the need for stricter privacy guarantees beyond output-level evaluations.

\noindent\textbf{Forgotten Knowledge Recovery Phenomenon}.  
Recent studies~\cite{qi2024fine_tuning,gao2024meta,hu2024jogging} on safety-aligned generative models have observed the \emph{Forgotten Knowledge Recovery Phenomenon} (FRP), where supposedly removed harmful knowledge unexpectedly re-emerges through targeted relearning. This suggests that unlearned targets leave residual traces. Existing work~\cite{gao2024meta,hu2024jogging} attributes these residuals to semantic/knowledge overlaps between harmful and other concepts. We go beyond this insights and examine FRP across a broad range of approximate machine unlearning (MU) methods~\cite{1modelsparisty23jia,kurmanji2023towards,zhao2024what,huang2024unified} in closed-problem settings (with well-defined input/output spaces), and leverage FRP to uncover privacy risks. We find that \textbf{algorithm-induced residuals persist regardless of semantic correlations}, which poses measurable membership privacy threats. %Some previous work \cite{chundawat2023zero} uses FRP speed to measure how closely an AMU matches a retrained model, but it assumes full access to training data, which is a impractical for attacks. However, these works assume full access to training data, which is impractical for real-world attacks. In contrast, our ReA attack requires only an inferred dataset to precisely capture the links between the FPR and membership privacy .

\section{Conclusion}
\label{sec:conclusion}
\vspace{-0.1cm}
%This paper exposes pervasive privacy vulnerabilities in approximate unlearning algorithms induced by implicit residuals. By exploiting these residuals, we propose ReA, a novel membership inference attack (MIA) to breach unlearned data privacy. ReA extends targets beyond unlearned samples to unlearned classes and concepts, introducing a broader privacy threat. Compared to current attacks (MIA-UP) relying on output-space residuals, ReA achieves up to $45\%$ higher accuracy. In response, we propose \emph{Orthogonal Unlearning \& Replay} (OUR), a framework that systematically scrubbing residuals to mitigate privacy leakage. OUR generalizes across class-wise, sample-wise, and concept unlearning, achieving near-retrained unlearning efficacy while reducing privacy attack accuracy nearly random guess.
This paper reveals pervasive privacy vulnerabilities in approximate unlearning algorithms induced by implicit residuals. To exploit these residuals, we introduce Reminiscence Attack (ReA), a novel membership inference attack (MIA) that compromises unlearned data privacy. ReA extends its targets beyond unlearned samples to unlearned classes and concepts, posing a broader privacy threat. Compared to existing attacks like MIA-UP~\cite{kurmanji2023towards} relying on output-space residuals, ReA achieves up to $1.90\times$ higher attack accuracy. To mitigate this risk, we propose Orthogonal Unlearning \& Replay (OUR), a framework that systematically scrubbing residuals to reduce privacy leakage. OUR generalizes across class-wise, sample-wise, and concept unlearning, achieving near-retrained unlearning efficacy while lowering privacy attack accuracy to near-random levels.

\section*{Acknowledgment}
This work was supported by the National Natural Science Foundation of China (Grant No: 92270123, 62372122, and 62372130), and the Research Grants Council, Hong Kong SAR, China (Grant No: 15210023, 15224124, 25207224).

\bibliographystyle{plain}
\bibliography{reference}

\newpage
\clearpage
\appendix
\appendix
\section*{Appendix}
\label{sec:appendix}
\section{Algorithms}
\label{app:algorithm}
In this section, we provide the pseudo-codes for our proposed privacy attack, the Reminiscence Attack (ReA) (described in Section~\ref{sec:attack}), and the unlearning framework Orthogonal Unlearning \& Replay (OUR) (defined in Section~\ref{sec:optimized_unlearning}).

\begin{algorithm}
	\caption{Class-wise \emph{Reminiscence Attack} (ReA)}
	\label{al:class_rea}
	\small
	\begin{algorithmic}[1]
		\STATE \textbf{Require}:  OOD dataset $\mathcal{D}^\text{ood}$, reference OOD set $\mathcal{D}^\prime$, assigned unlearned label $y_u$, loss function $\mathcal{L}$
		\STATE \textbf{Parameters}: Max iterations $\text{Idx}_\text{max}$, learning rates $\{\text{lr}_j\}_{j=1}^J$
		
		\IF{Black-Box Scenario}
		\STATE $\theta'_0 \leftarrow \text{ModelExtract}(F(\cdot;\theta^u))$ \COMMENT{\textcolor{gray}{Model extraction}}
		\STATE $F_\text{ini} \gets \text{Copy}(F(\cdot;\theta'_0))$ \COMMENT{\textcolor{gray}{Save initial model}}
		\ELSIF{White-Box Scenario}
		\STATE $\theta'_0 \gets \theta^u$ \COMMENT{\textcolor{gray}{Direct access}}
		\ENDIF
		
		\FOR{$\text{lr}_j \in \{\text{lr}_j\}_{j=1}^J$}
		\STATE Initialize $\theta' \gets \theta'_0$
		\FOR{$i = 1$ \TO $\text{Idx}_\text{max}$}
		\STATE Compute loss:
		{\small
			$L = \frac{1}{|\mathcal{D}^\text{ood}|} \sum_{x\in \mathcal{D}^\text{ood}} \mathcal{L}(F(x;\theta'), y_u) 
			+ \frac{1}{|\mathcal{D}^\prime|} \sum_{x\in \mathcal{D}^\prime} \mathcal{L}(F(x;\theta'), F_\text{ini}(x))
			$}
		\STATE $\theta' \gets \theta' - \text{lr}_j \cdot \nabla_{\theta'} L$ \COMMENT{\textcolor{gray}{SGD update}}
		\IF{$\text{Acc}_{\mathcal{D}^\text{ood}}(F(\cdot;\theta')) > 0.9$}
		\STATE Record $\text{Idx}_\text{r}(\mathcal{D}^\text{ood}, \text{lr}_j) \gets i$
		\STATE \textbf{break}
		\ENDIF
		\ENDFOR
		\ENDFOR
		
		\STATE Compute confidence score:
		
		~~~~{\small
			$\mathcal{A}'(\mathcal{D}^\text{ood}) = 1 - \frac{1}{J \cdot \text{Idx}_\text{max}} \sum_{j=1}^J \text{Idx}_\text{r}(\mathcal{D}^\text{ood}, \text{lr}_j)$~Equation~\ref{eq:aggre_conf_class}
		}
		\STATE \textbf{Output}: Confidence score $\mathcal{A}'(\mathcal{D}^\text{ood})$
	\end{algorithmic}
\end{algorithm}

\begin{algorithm}
	\caption{Sample-wise \emph{Reminiscence Attack} (ReA)}
	\label{al:sample_rea}
	\small
	\begin{algorithmic}[1]
		\STATE \textbf{Require}:  Inference dataset $\mathcal{D}_\text{infer}$, victim model $F(\cdot;\theta^u)$ \COMMENT{\textcolor{gray}{i.e., the unlearned model}}
		\STATE \textbf{Parameters}: Pseudo retain set size $N_r$, training epochs $\text{EPOCHS}=5$, learning rate $\text{lr}$
		
		\STATE Obtain confidence scores: $\{s\} \gets \mathcal{A}'(\mathcal{D}_\text{infer})$ \COMMENT{\textcolor{gray}{Using MIA-LiRA method}}
		\STATE Select top-$N_r$ indices $\{\text{ind}_i\}_{i=1}^{N_r}$ from $\{s\}$ based on confidence
		\STATE Construct pseudo retain set: $\mathcal{D}_r' \gets \mathcal{D}_\text{infer}[\{\text{ind}_i\}_{i=1}^{N_r}]$ \COMMENT{\textcolor{gray}{Subset selection}}
		
		\FOR{$e = 1$ \TO $\text{EPOCHS}$}
		\STATE Compute loss:
		
		{\small
			$L = \frac{1}{|\mathcal{D}_r'|} \sum_{(x,y)\in \mathcal{D}_r'} \mathcal{L}(F(x;\theta'), y)$
		}
		\STATE $\theta' \gets \theta' - \text{lr} \cdot \nabla_{\theta'} L$ \COMMENT{\textcolor{gray}{SGD update}}
		\ENDFOR
		\STATE \textbf{Output}: Updated confidence scores $\{\mathcal{A}'(F(x; \theta'))\}_{x \in \mathcal{D}_\text{infer}}$
	\end{algorithmic}
\end{algorithm}

\begin{algorithm}
	\caption{\emph{Orthogonal Unlearning \& Replay} (OUR) Framework}
	\label{al:over-replay}
	\small
	\begin{algorithmic}[1]
		\STATE \textbf{Require}: Original model $F(\cdot;\theta)$, full dataset $\mathcal{D}$, unlearning set $\mathcal{D}_u$
		\STATE \textbf{Parameters}: Phase epochs $e_1$, $e_2$, norm threshold $\Delta_\text{thr}$, learning rate $\text{lr}$
		
		\STATE Compute remaining set: $\mathcal{D}_r \gets \mathcal{D} \setminus \mathcal{D}_u$
		\STATE Store initial parameters: $\theta^\text{ini} \gets \theta = [\theta_1, \cdots]$
	\end{algorithmic}
	\textbf{Phase 1: The Orthogonal Unlearning}
	\begin{algorithmic}[1]
		\FOR{$e = 1$ \TO $e_1$}
		\STATE Compute orthogonal loss:
		
		{\small
			$L_\text{orth} = \sum_{(x, y)\in\mathcal{D}_u} \sum_{l=1}^k \|F_l(x; \theta) - F_l(x; \theta^0)\|_2^2$}
		
		\STATE $\theta \gets \theta - \text{lr} \cdot \nabla_\theta L_\text{orth}$ \COMMENT{\textcolor{gray}{Orthogonality Update}}
		\STATE Compute the maximum parameter change: 
		
		~~~$\Delta_\text{max} \gets\max_{i\in |\theta|}{\frac{\|\theta_i - \theta_i^\text{ini}\|_2}{|\theta_i|}}$
		\IF{$\Delta_\text{max} > \Delta_\text{thr}$}
		\STATE \textbf{break} \COMMENT{\textcolor{gray}{Early stopping}}
		\ENDIF
		\ENDFOR
	\end{algorithmic}
	\textbf{Phase 2: The Replay Phase}
	\begin{algorithmic}[1]
		\FOR{$e = 1$ \TO $e_2$}
		\STATE Compute loss:
		
		{\small 
			$L_ = \frac{1}{|\mathcal{D}_r|} \sum_{(x,y)\in \mathcal{D}_r} \mathcal{L}(F(x;\theta), y)$}		
		\STATE $\theta \gets \theta - \text{lr} \cdot \nabla_\theta L$ \COMMENT{\textcolor{gray}{Retain knowledge}}
		\ENDFOR
		\STATE \textbf{Output}: Unlearned model $F(\cdot;\theta^u)$ where $\theta^u \gets \theta$
	\end{algorithmic}
\end{algorithm}

\section{The Proportion of Stable Neurons after Orthogonal Unlearning}
\label{app:our_stable_neurons}
Orthogonal unlearning induces rapid utility degradation by excluding the retained dataset, yet model functionality recovers efficiently in the replay phase due to minimal parameter perturbations. To validate this, we empirically analyze neuron stability in a ViT model trained on CIFAR-20 after class-wise unlearning.

We define the change for each parameter as follows. Since parameters vary in size, we define the change of parameter $\theta_i$ as:

\begin{equation}
	\small
	\Delta(\theta_i) = {\frac{\|\theta_i - \theta_i^\text{ini}\|_2}{|\theta_i|}},
\end{equation} 
where $\theta_i^\text{ini}$ represents the parameter's state before unlearning. To account for differences in parameter dimensions, we normalize the change by its size $|\theta_i|$ for better observation. The maximum parameter change is then defined as:
\begin{equation}
	\small
	\Delta_\text{max} = \max_{i\in |\theta|}{\frac{\|\theta_i - \theta_i^\text{ini}\|_2}{|\theta_i|}}
\end{equation} 

\begin{figure}
	\centering
	\includegraphics[width=0.6\columnwidth]{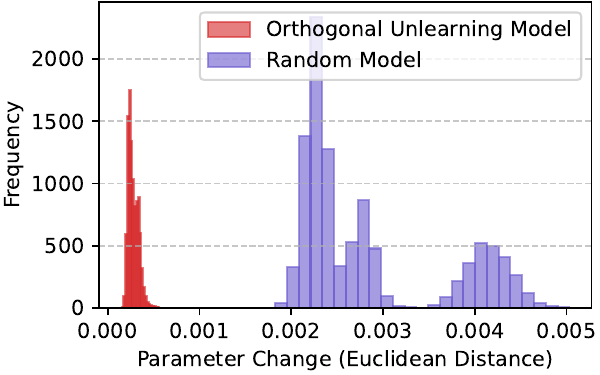}
	\caption{Histograms of Parameter Changes.}
	\label{fig:histo_stable}
\end{figure}

\begin{figure}
	\centering
	\includegraphics[width=0.6\columnwidth]{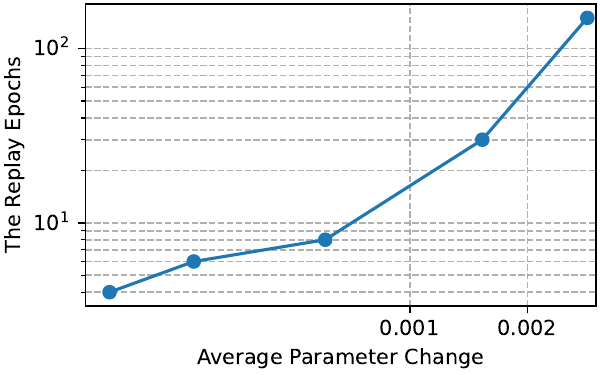}
	\caption{Parameter Changes versus The replay speed measured by epochs.}
	\label{fig:changes_vs_replay_speed}
\end{figure}

Figure~\ref{fig:histo_stable} reports that parameter changes after unlearning remain highly localized, with magnitudes concentrated below $0.0005$, which is three times smaller than the most minor changes observed in random models. Further analysis in Figure~\ref{fig:changes_vs_replay_speed} confirms that smaller parameter changes correlate strongly with faster utility recovery during replay, which requires fewer training epochs. These results demonstrate that orthogonal unlearning preserves critical neural patterns despite utility drops, enabling efficient restoration through targeted replay rather than full retraining. 

\section{Supplementary Visualized Loss Landscape}
\label{app:loss_landscape}
\begin{figure}
	\centering
	\hspace{-0.3cm}\subfloat[Class-wise Unlearn (RUM~\cite{zhao2024what})]{
		\label{fig:class_landscape-cifar100}
		\includegraphics[width=0.468\columnwidth]{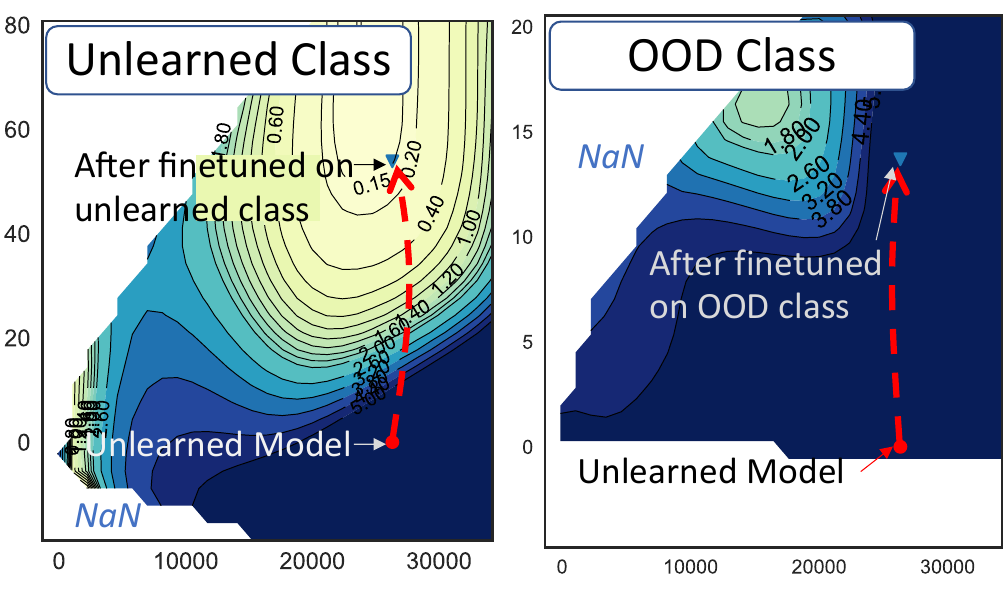}}
	\subfloat[Sample-wise Unlearn (GA~\cite{thudi2022unrolling})]{
		\label{fig:sample_landscape-cifar10}
		\includegraphics[width=0.468\columnwidth]{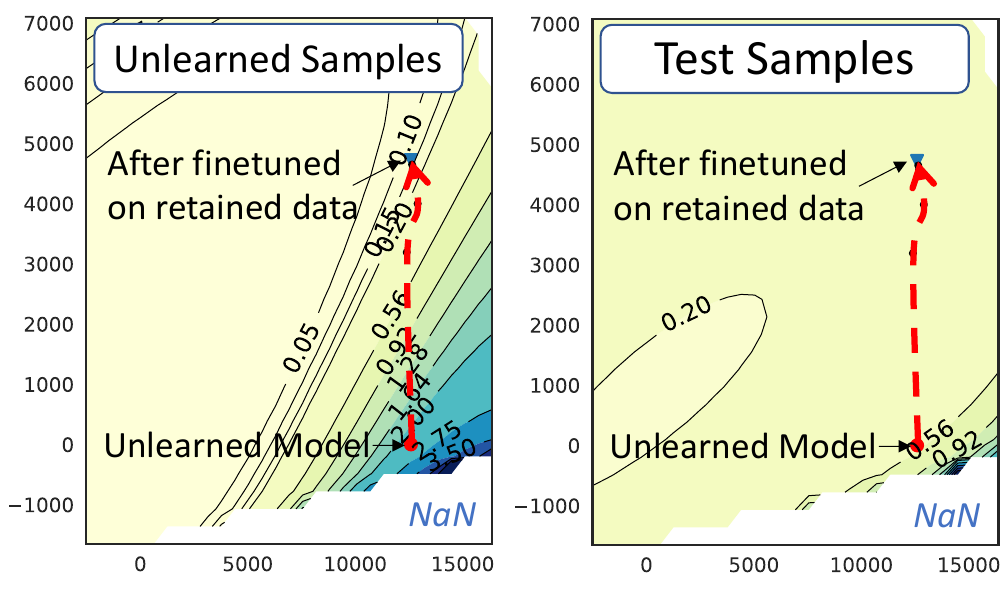}}
	\vspace{-0.25cm}
	\caption{Visualized loss landscapes of unlearned data and non-training data around unlearned models in CIFAR100 and CIFAR10 experiments.}%, visualized by the method in~\cite{zheng2025spurious}.}
\vspace{-0.35cm}
\label{app:fig:landscape}
\end{figure}

We extend Figure~\ref{fig:landscape} to CIFAR-10/100-ResNet18 in Figure~\ref{app:fig:landscape}, confirming that residuals, i.e., discrepancies between unlearned and non-training data, consistently appear across architectures. These are reflected in fine-tuning trajectories (dashed lines): in class-wise unlearning (Figure~\ref{fig:class_landscape-cifar100}), the loss of unlearned classes converges to sharp minima while OOD classes remain unconverged; in sample-wise unlearning (Figure~\ref{fig:sample_landscape-cifar10}), unlearned samples exhibit sharp loss drops while test losses stay stable. These differences expose membership privacy.

\section{Experimental Setups and Approximate Unlearning Benchmarks}
\label{app:ex_setups}

\subsection{Setups}
\label{app:ex_setups_sub}
\noindent\textbf{ReA Setups}. 
For class-wise ReA attacks, the ratio of reference data for logits constraints to inferred data is set to $14:1$ for CIFAR20 and $6:1$ for CIFAR100 during reminiscence. The maximum number of training iterations $\text{Idx}_{\text{max}}$ is set to $75$. We employ cross entropy loss $\mathcal{L}$ during its reminiscence process, with SGD with momentum ($0.9$) and weight decay $5$e$^{-4}$. In sample-wise ReA, the size of ``pseudo'' retained dataset $\mathcal{N}_r$ is $20,\!000$.

\noindent\textbf{OUR Unlearning Method Setups}.  
For the implementation of OUR, we adopt the following hyper-parameter configurations. The training process consists of distinct phases tailored for different architectures.  

For ResNet18 experiments, the orthogonal unlearning and replay phases each span $8$ epochs, with learning rates of $0.0018$ and $0.005$, respectively; a $0.5$ decay is applied at epoch $3$ during replay phases. For ViT experiments, the unlearning phase lasts $7$ epochs. The replay phase runs for $7$ epochs for class-wise unlearning or $11$ epochs for sample-wise unlearning, with learning rates of $0.0008$ and $0.0004$, and decay factors of $0.5$ and $0.2$ at epochs $4$ and $6$. The ablation study on the number of epochs in OUR is provided in Appendix~\ref{app:exp_abl_e1e2}. Besides, the regularization component involves an $l_1$ regularization factor of $1$e$^{-5}$ to enhance model sparsity~\cite{1modelsparisty23jia}. 

\noindent\textbf{Implementation Details}. 
All classification experiments are conducted using PyTorch on 2$\times$RTX 4090 GPUs, and diffusion model experiments are performed on 4$\times$RTX 4090 GPUs.

\subsection{Approximate Machine Unlearning Benchmarks}
\label{app:mu_benchmark}
\begin{itemize}
	\item {\bf Fine-tuning (FT)}~\cite{3machineunlearning23warnecke,1modelsparisty23jia}. It leverages the phenomenon of catastrophic forgetting to continue training the pre-trained model $F(.;\theta)$ on the remaining dataset $\mathcal{D}_r$ for a limited number of iterations, to forget the unlearned dataset $\mathcal{D}_u$ which is no longer trained upon. 
	
	\item {\bf Gradient Ascent (GA)}~\cite{thudi2022unrolling,amnesiac21graves}. This approach seeks to achieve unlearning by reversing the gradient descent process on $\mathcal{D}_u$, thereby erasing its optimization traces in the model $F(.;\theta)$. 
	
	\item {\bf Random Label (RL)}~\cite{amnesiac21graves, 6can23chundawat}. It assigns random labels for $\mathcal{D}_u$ to erase the model $F(.;\theta)$'s memory of them.
	
	\item {\bf Influence Unlearning(IU)}~\cite{1modelsparisty23jia}. It utilizes the woodfisher method. It estimates the influence of $\mathcal{D}_u$ on $F(.;\theta)$ and designs perturbation strategies to erase this influence from the parameters $\theta$.
	
	\item {\bf FisherForgetting (FF)}~\cite{2eternal20golatkar}. FF perturbs $\theta$ with additive Gaussian noise, where the covariance is derived from the fourth root of the Fisher Information matrix over $\mathcal{D}_r$. While theoretically rigorous, its reliance on Fisher matrix inversion limits parallel efficiency and increases computational overhead compared to gradient-based methods.
	
	\item {\bf Boundary-based Unlearning (BU)}~\cite{chen2023boundary}. BU shifts the decision boundary of $F(.;\theta)$ to mimic a retrained model’s behavior, thereby bypassing parameter-space optimization.
\end{itemize}

Additionally, there are five optimized unlearning frameworks.
\begin{itemize}	
	\item {\bf $l1$ sparsity}~\cite{1modelsparisty23jia}. This framework bridges approximate and exact unlearning by pruning non-critical weights. Sparsity reduces the parameter space, thus it narrows the gap between approximate and ideal unlearning outcomes while maintaining efficiency.
	
	\item {\bf SalUn}~\cite{fan2024salun}. It introduces weight saliency to focus unlearning efforts on critical parameters. Analogous to input saliency in explainable AI, it prioritizes weights with high influence on $\mathcal{D}_u$.
	
	\item {\bf SCRUB}~\cite{kurmanji2023towards}. It uses a teacher-student architecture where the student selectively disregards the teacher’s knowledge about $\mathcal{D}_u$. This ``unlearning-by-disobedience'' approach scales without restrictive assumptions.
	
	\item {\bf SFRon}~\cite{huang2024unified}. It unifies gradient-based MU by decomposing updates into three components: forgetting gradient ascent, retaining gradient descent, and saliency-guided weighting. It further incorporates a Hessian-aware manifold geometry to align unlearning trajectories with the output probability space to balancing utility performance and forgetting efficacy.
	
	\item {\bf Refined-Unlearning Meta-algorithm (RUM)}~\cite{zhao2024what}. It refines the forget set $\mathcal{D}_u$ into homogeneous subsets and applies specialized unlearning strategies to each. Its meta-algorithm orchestrates existing methods to comprehensively unlearn $\mathcal{D}_u$.
\end{itemize}

\section{Complementary Experimental Results}
\subsection{Ablation Study of Layers $\{l\}_k$ Seleted in OUR}
\label{app:exp_abl_layerchoice}
To determine the optimal layer configuration for orthogonal unlearning (OUR), we conduct an ablation study evaluating three distinct layer-selection strategies for ${l}_k$: using output layers from the first three transformer blocks (\textbf{First 3}), the last three blocks (\textbf{Last 3}), and a distributed set comprising the first, middle, and final blocks (\textbf{Span 3}). Experiments on CIFAR-20 evaluate unlearning performance through ToW of Unlearning and resistance against relearning attacks through ReA Accuracy, where values approaching $1.0$ and $50\%$ respectively indicate optimal outcomes. 

\begin{figure}
	\centering
	\includegraphics[width=0.76\columnwidth]{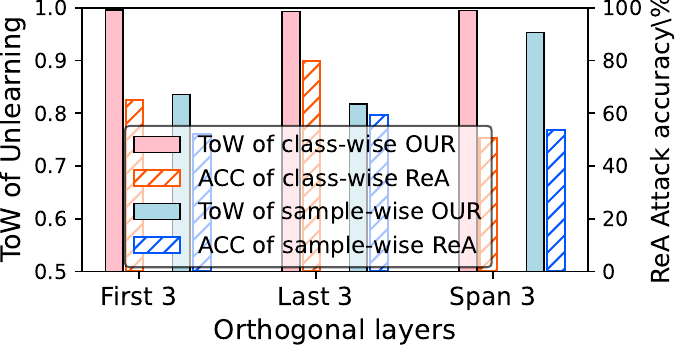}
	\caption{OUR performance with different $\{l\}_k$ settings: first 3 / last 3 / span 3 denote the output layers of the first 3 blocks, last 3 blocks, and first, middle, and last blocks, respectively.}
	\label{app:fig_l_k}
\end{figure}

As shown in Figure~\ref{app:fig_l_k}, the Span 3 configuration achieves the most favorable balance across both evaluation metrics for class-wise and sample-wise OUR implementations, which attains near-optimal ToW while maintaining ReA accuracy closest to random-guess performance.

\subsection{Ablation Study of Number of Epochs in OUR}
\label{app:exp_abl_e1e2}
To optimize the training efficiency and unlearning effectiveness of OUR unlearning, we conduct an ablation study examining epoch configurations for its two-phase training: orthogonal unlearning ($e1$) and replay ($e2$). We evaluate three critical metrics across five repeated trials on CIFAR20-ViT, CIFAR10-ResNet18, and CIFAR100-ResNet18: unlearning performance (ToW of Unlearning), and resistance against relearning attacks (ReA Attack Accuracy).

\begin{figure}
	\centering
	\subfloat[Class-wise AMU(ViT)]{
		\includegraphics[width=0.468\columnwidth]{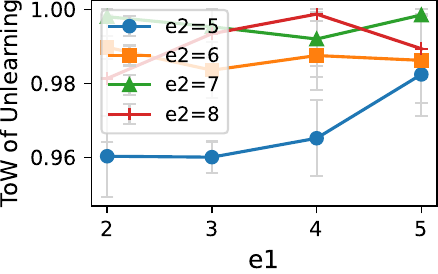}}
	\subfloat[Class-wise AMU(ViT)]{
		\includegraphics[width=0.468\columnwidth]{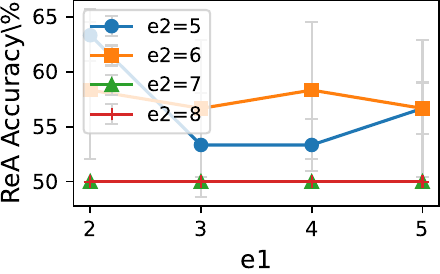}}\\
	\subfloat[Sample-wise AMU(ViT)]{
		\includegraphics[width=0.468\columnwidth]{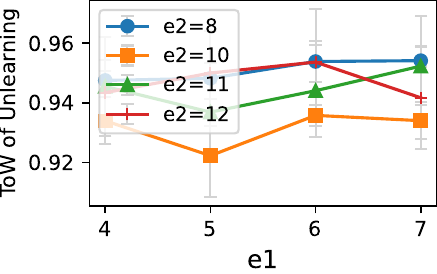}}
	\subfloat[Sample-wise AMU(ViT)]{
		\includegraphics[width=0.468\columnwidth]{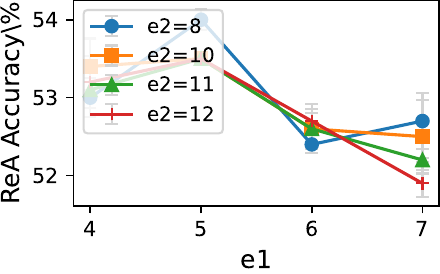}}\\
	\subfloat[Class-wise AMU(ResNet18)]{
		\includegraphics[width=0.468\columnwidth]{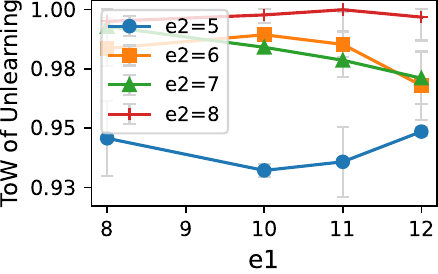}}
	\subfloat[Class-wise AMU(ResNet18)]{
		\includegraphics[width=0.468\columnwidth]{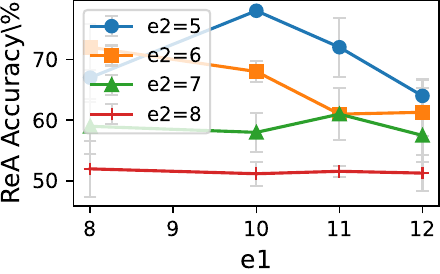}}\\
		\subfloat[Sample-wise AMU(ResNet18)]{
			\includegraphics[width=0.468\columnwidth]{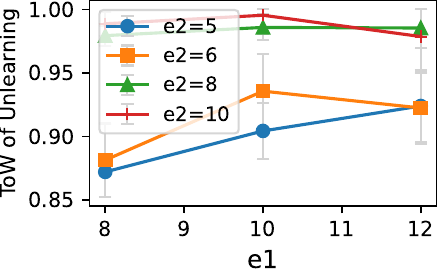}}
	\subfloat[Sample-wise AMU(ResNet18)]{
		\includegraphics[width=0.468\columnwidth]{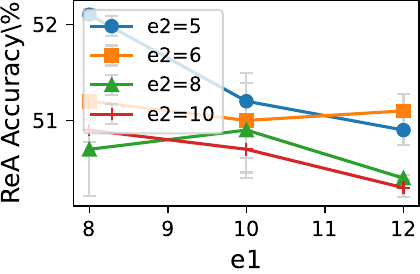}}
	\caption{Ablation study of $e_1$ and $e_2$ in OUR. Gray vertical bars indicate error bars. Class-wise ReA is evaluated 10 times for both unlearn and OOD classes per experiment. Results are averaged over 5 independent trials.}
\label{app:fig_e1e2}
\end{figure}

As shown in Figure~\ref{app:fig_e1e2}, optimal configurations emerge for each setting: CIFAR20-ViT requires 2 $e1$ and 7 $e2 $ for sample-wise and 7 $e1$ and 11 $e2 $ for class-wise OUR; CIFAR10-ResNet18 requires 8 $e1$ and 8 $e2 $ for sample-wise OUR; CIFAR100-ResNet18 requires 8 $e1$ and 8 $e2 $ for class-wise OUR. These configurations simultaneously achieve near-optimal ToW ($\approx1.0$) and strong ReA resistance while minimizing RTE.

\subsection{Ablation Study of  ReA Learning Rate Set}
\label{subsec:rea_lr}
The learning rate (lr) during the reminiscence process largely impacts class-wise ReA efficacy, as it directly affects the scores ($\mathcal{A}^\prime$ in Equation~\ref{eq:class_mia}) for membership detection. In contrast, sample-wise ReA remains robust to lr variations, requiring only a minimal lr of $0.1\times$ the training lr for model convergence. Here, we evaluate whether our multi-lr aggregation strategy (Section~\ref{subsec:rml}) enhances robustness to this parameter in class-wise ReA.

\begin{figure}
	\centering
	\includegraphics[width=1\columnwidth]{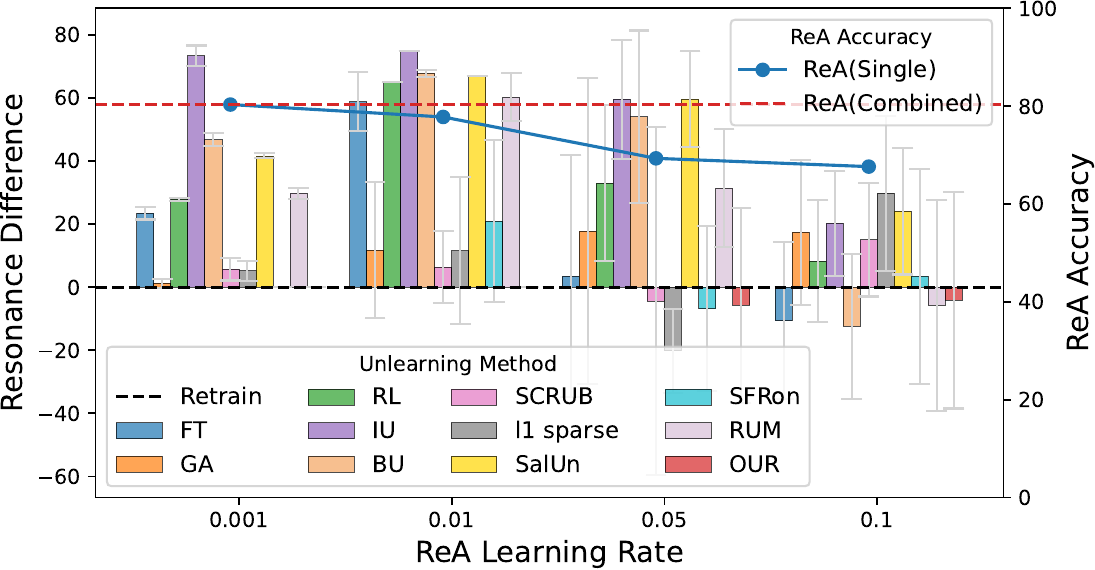}
	\caption{Resonance difference (Idx\textsubscript{r} of OOD classes minus that of unlearned classes) and ReA attack accuracy across learning rates in ReA. Gray bars represent error bars.}
	\label{fig:lr_vs_resonace_diff}
\end{figure}

Figure~\ref{fig:lr_vs_resonace_diff} compares resonance differences (the difference of resonance index between OOD and unlearned classes, $\Delta$Idx\textsubscript{r}) for all MU methods at different lr on CIFAR-20, which is presented as bars. The left y-axis shows resonance differences $\Delta$Idx\textsubscript{r} (positive values indicate faster convergence for unlearned classes), averaged over 10 trials. The right y-axis contrasts ReA privacy attack accuracy under single-lr (blue lines) and multi-lr aggregation (red dashed line) strategies. \noindent\textbf{Key observations.} First, single lr$=0.001$ maximizes ReA effectiveness for most MU methods except GA~\cite{thudi2022unrolling} and its variant, SFRon~\cite{huang2024unified}, which indicates that optimal lr ranges are dependent on MU methods. Second, ReA's performance collapses at lr$=0.1$. Third, the multi-lr aggregation strategy achieves superior results without method-specific tuning. Notably, our OUR method exhibits near-zero resonance difference, demonstrating inherent resistance to ReA attacks.

\subsection{Ablation Study of ReA Convergence Threshold}
\label{app:rea_threshold}
\begin{figure}
	\centering
	\includegraphics[width=0.71\columnwidth]{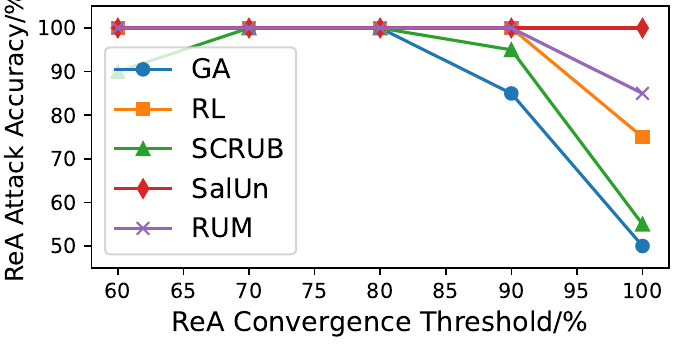}
	\caption{ReA Attack Accuracy versus Convergence Threshold}
	\label{app:fig_convergence_thre}
\end{figure}

To evaluate how convergence threshold affects ReA performance, we conduct an ablation study on this key parameter. The convergence threshold represents the prediction accuracy threshold in class-wise ReA to determine the resonance index. Experiments perform on CIFAR-20 and evaluate the ReA performance targeting five selected unlearning benchmarks: GA, RL, SCRUB, SalUn, and RUM. The results in Figure~\ref{app:fig_convergence_thre} confirm that thresholds between $70\%$ and $80\%$  yield optimal ReA attack accuracy across all evaluated unlearning methods.

\subsection{The Complete ROC Analysis of Privacy Attacks in Section~\ref{subsec:ex_overall}}
\label{app:roc_full}

\begin{figure}
	\centering
	\subfloat[Retrain]{
		\includegraphics[width=0.31\columnwidth]{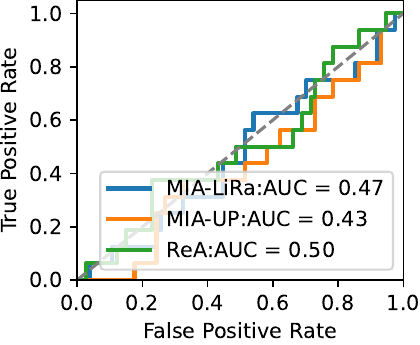}}
	\subfloat[FT]{
		\includegraphics[width=0.31\columnwidth]{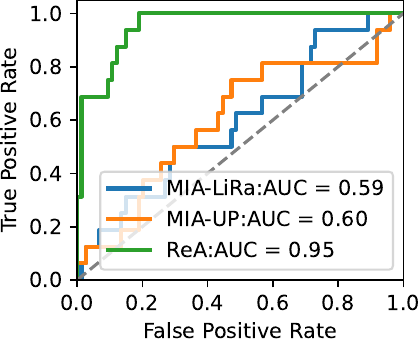}}
	\subfloat[GA]{
		\includegraphics[width=0.31\columnwidth]{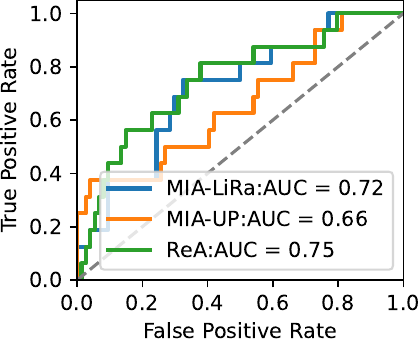}}\\
	\subfloat[RL]{
		\includegraphics[width=0.31\columnwidth]{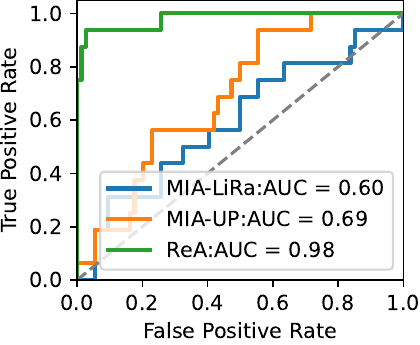}}
	\subfloat[IU]{
		\includegraphics[width=0.31\columnwidth]{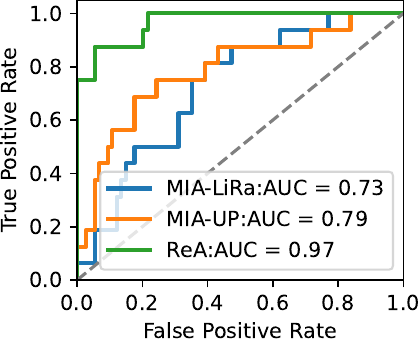}}
	\subfloat[BU]{
		\includegraphics[width=0.31\columnwidth]{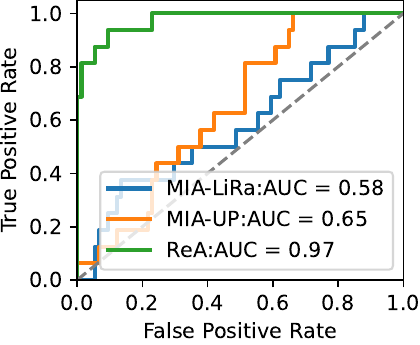}}\\
	\subfloat[SCRUB]{
		\includegraphics[width=0.31\columnwidth]{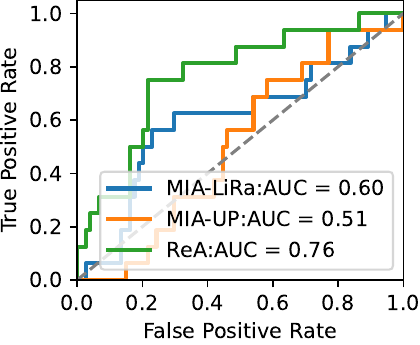}}
	\subfloat[$l1$ sparse]{
		\includegraphics[width=0.31\columnwidth]{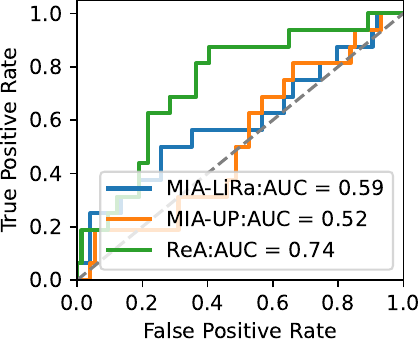}}
	\subfloat[SalUn]{
		\includegraphics[width=0.31\columnwidth]{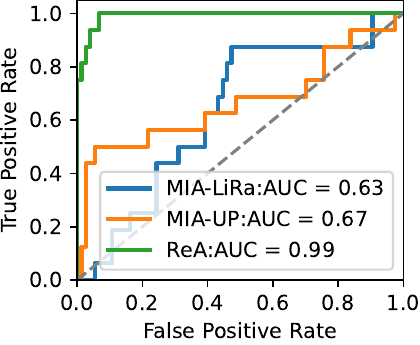}}\\
	\subfloat[SFRon]{
		\includegraphics[width=0.31\columnwidth]{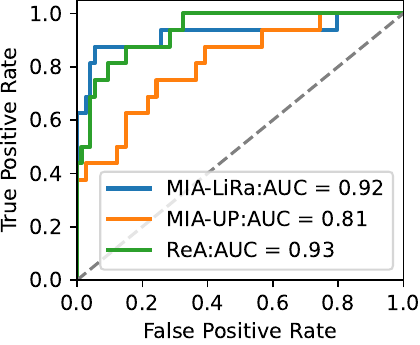}}
	\subfloat[RUM]{
		\includegraphics[width=0.31\columnwidth]{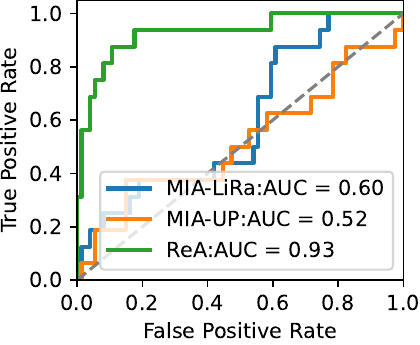}}
	\subfloat[OUR]{
		\includegraphics[width=0.31\columnwidth]{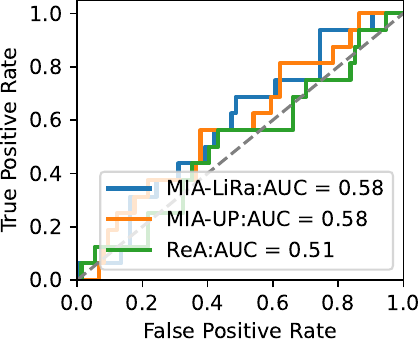}}
	\caption{ROC of class-wise MIA. The notable fluctuations result from having only 300 inferred objects in each figure.}
	\label{fig:app_auc_class_wise}
\end{figure}

\begin{figure}
	\centering
	\subfloat[Retrain]{
		\includegraphics[width=0.31\columnwidth]{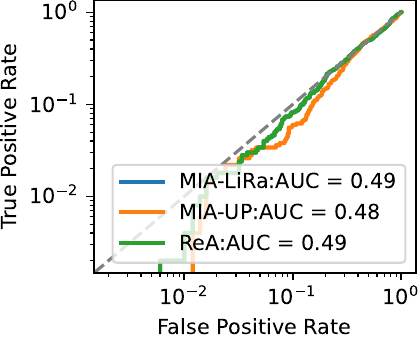}}
	\subfloat[FT]{
		\includegraphics[width=0.31\columnwidth]{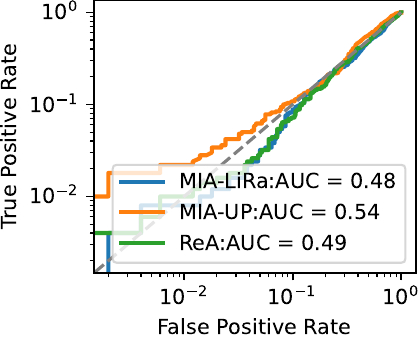}}
	\subfloat[GA]{
		\includegraphics[width=0.31\columnwidth]{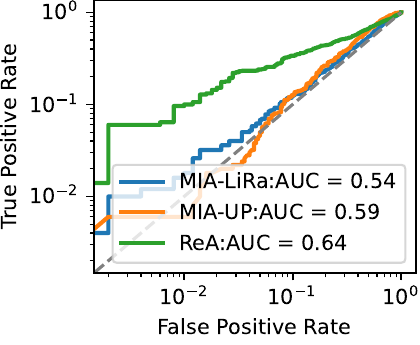}}\\
	\subfloat[RL]{
		\includegraphics[width=0.31\columnwidth]{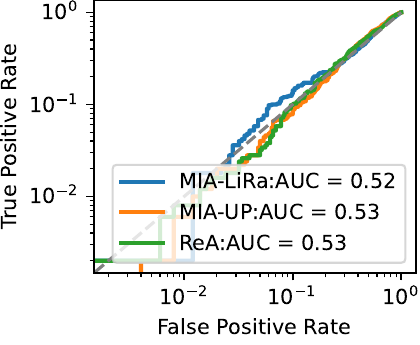}}
	\subfloat[IU]{
		\includegraphics[width=0.31\columnwidth]{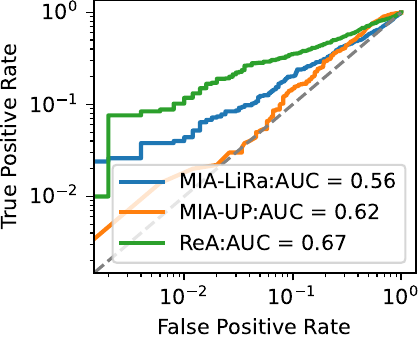}}
	\subfloat[FF]{
		\includegraphics[width=0.31\columnwidth]{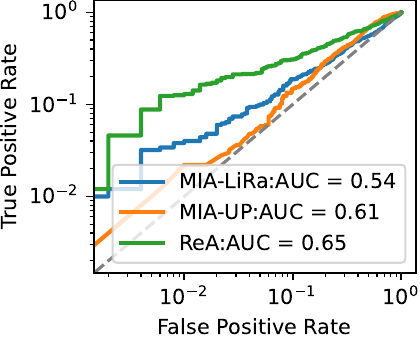}}\\
	\subfloat[SCRUB]{
		\includegraphics[width=0.31\columnwidth]{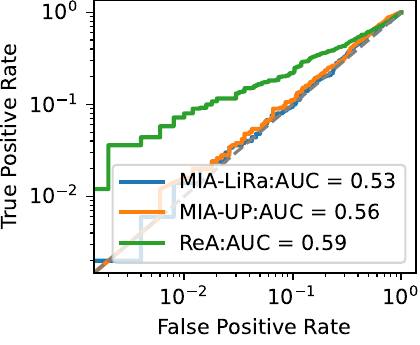}}
	\subfloat[$l1$ sparse]{
		\includegraphics[width=0.31\columnwidth]{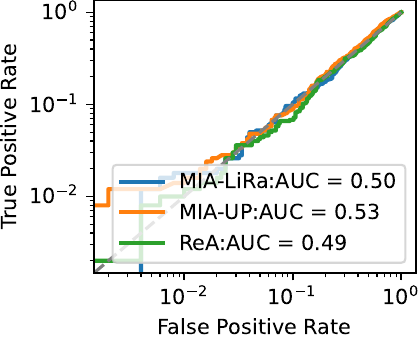}}
	\subfloat[SalUn]{
		\includegraphics[width=0.31\columnwidth]{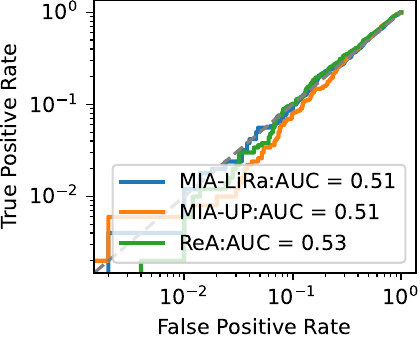}}\\
	\subfloat[SFRon]{
		\includegraphics[width=0.31\columnwidth]{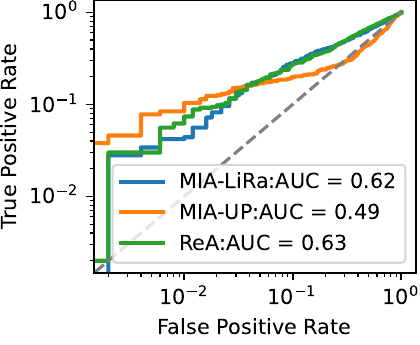}}
	\subfloat[RUM]{
		\includegraphics[width=0.31\columnwidth]{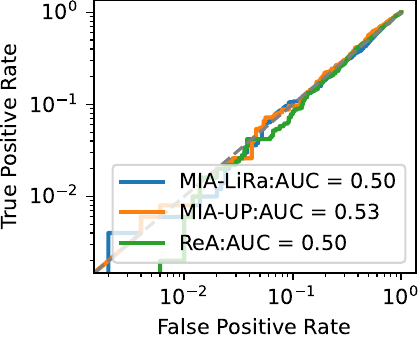}}
	\subfloat[OUR]{
		\includegraphics[width=0.31\columnwidth]{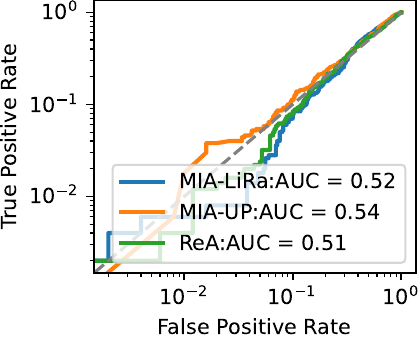}}
	\caption{ROC of sample-wise MIA.}
	\label{fig:app_auc_sample_wise}
\end{figure}

This section provides the full ROC analysis for machine unlearning benchmarks in CIFAR20 experiments. Figure~\ref{fig:app_auc_class_wise} shows the ROC figures of class-wise membership inference attacks (MIA). Figure~\ref{fig:app_auc_sample_wise} shows the ROC figures of sample-wise membership inference attacks (MIA).

\subsection{The Complete Representation Visualization Analysis of Experiments in Section~\ref{subsec:visual_reps}}
\label{app:representation_visual}
\begin{figure}
	\centering
	\subfloat[Retrain]{
		\includegraphics[width=0.3\columnwidth]{tsne_retrain.png}}~
	\subfloat[FT~\cite{amnesiac21graves}]{
		\includegraphics[width=0.3\columnwidth]{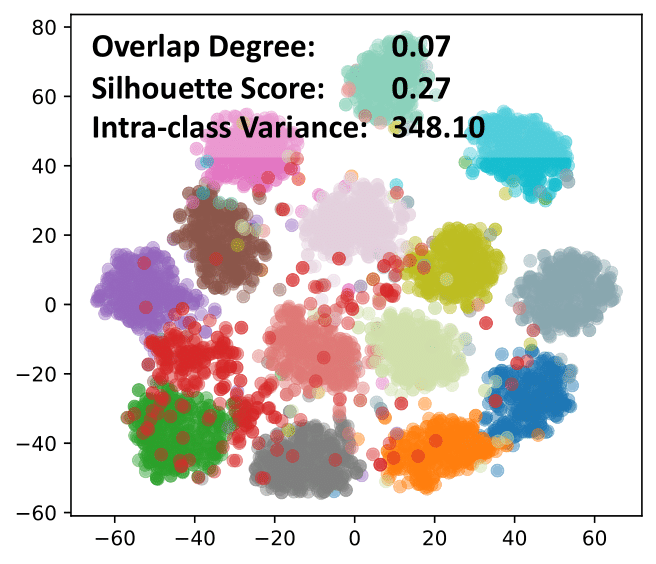}}~
	\subfloat[GA~\cite{3machineunlearning23warnecke}]{
		\includegraphics[width=0.3\columnwidth]{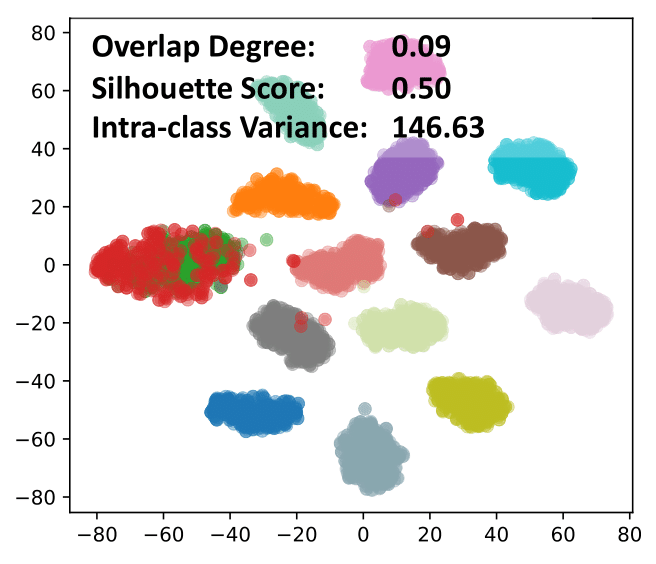}}\\
	\subfloat[RL~\cite{amnesiac21graves}]{
		\includegraphics[width=0.3\columnwidth]{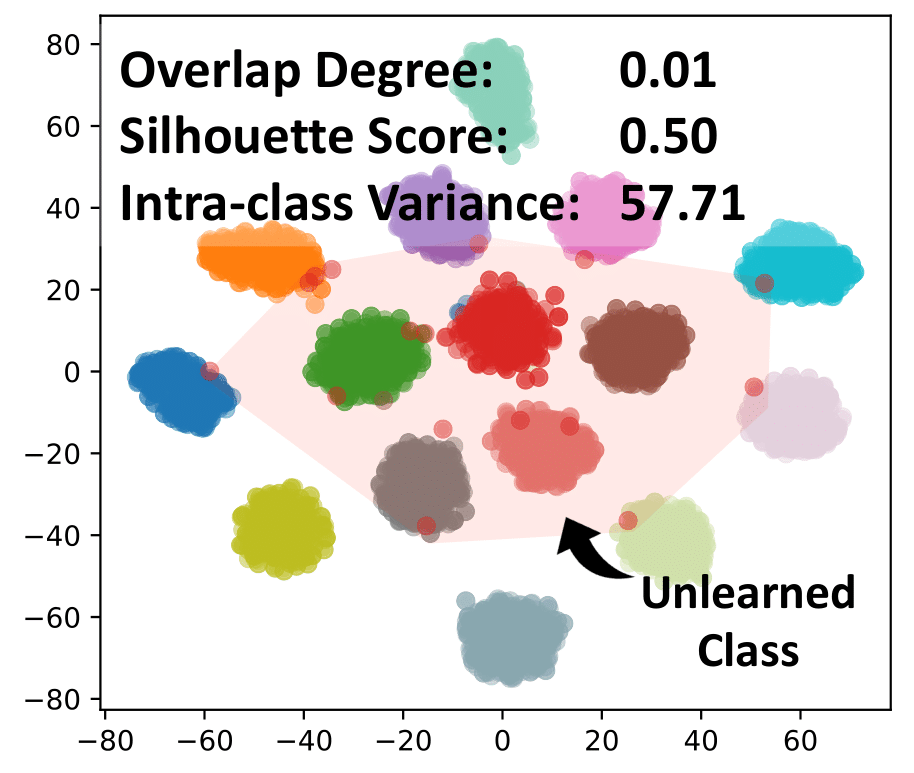}}~
	\subfloat[IU~\cite{1modelsparisty23jia}]{
		\includegraphics[width=0.3\columnwidth]{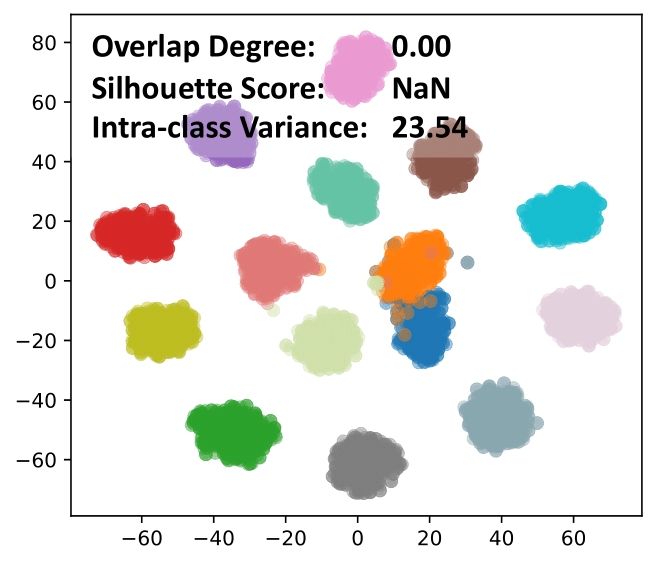}}~
	\subfloat[BU~\cite{chen2023boundary}]{
		\includegraphics[width=0.3\columnwidth]{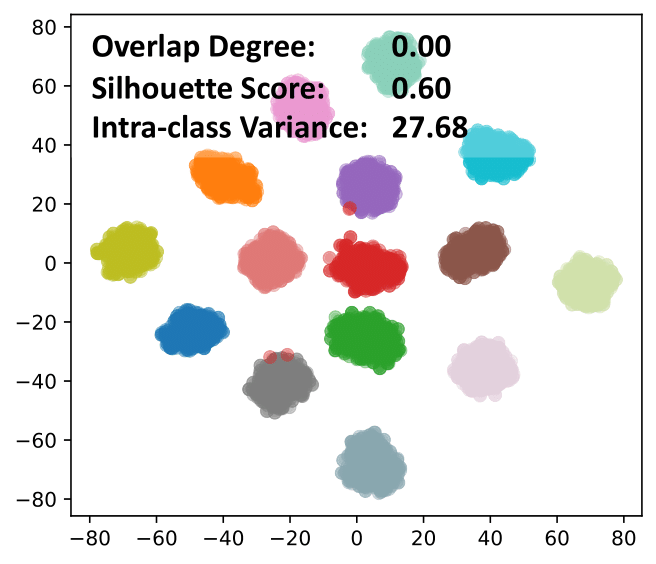}}\\
	\subfloat[SCRUB~\cite{kurmanji2023towards}]{
		\includegraphics[width=0.3\columnwidth]{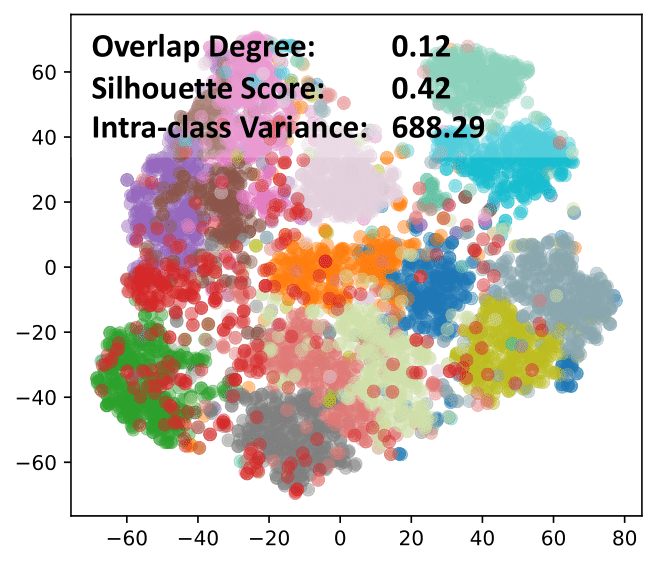}}~
	\subfloat[$l1$ sparse~\cite{1modelsparisty23jia}]{
		\includegraphics[width=0.3\columnwidth]{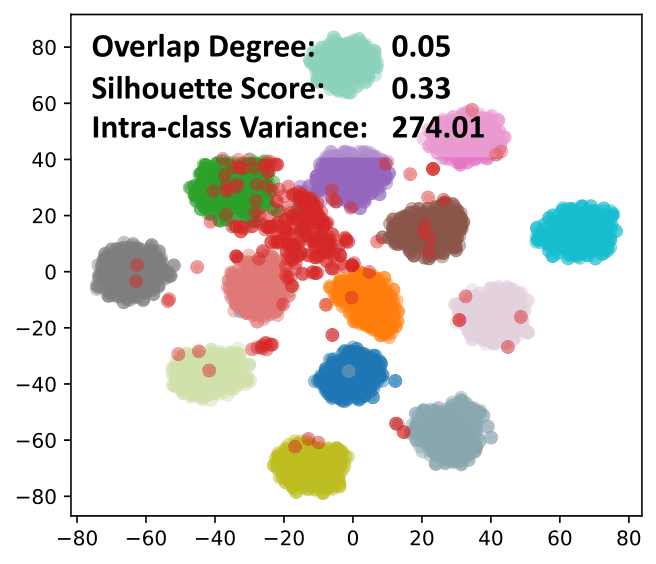}}~
	\subfloat[Salun~\cite{fan2024salun}]{
		\includegraphics[width=0.3\columnwidth]{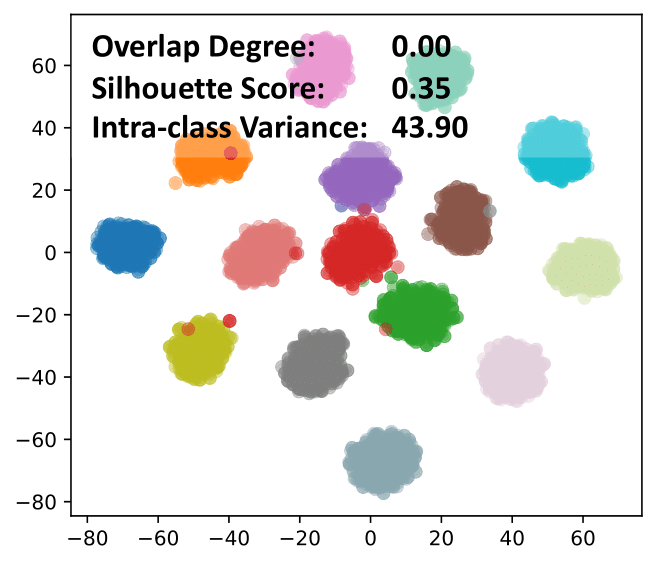}}\\
	\subfloat[SFRon~\cite{huang2024unified}]{
		\includegraphics[width=0.3\columnwidth]{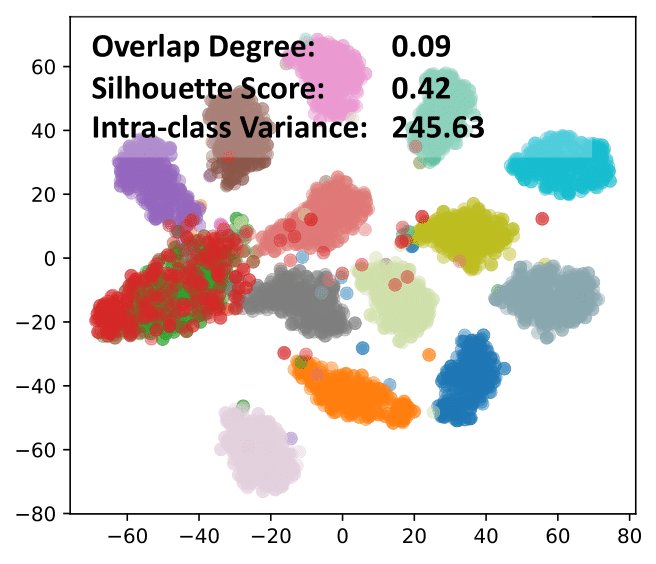}}~
	\subfloat[RUM~\cite{zhao2024what}]{
		\includegraphics[width=0.3\columnwidth]{tsne_rum.png}}~
	\subfloat[OUR]{
		\includegraphics[width=0.3\columnwidth]{tsne_our.png}}
	% \vspace{-0.13cm}
	\caption{Representation space visualization of the class-wise unlearned model in CIFAR20 experiments.}
	\label{fig:app_tsne_ex}
\end{figure}

This appendix provides additional visualizations of lower-dimensional embeddings for  Figure~\ref{fig:tsne_ex} in Section~\ref{subsec:visual_reps}. All visualization results are presented in Figure~\ref{fig:app_tsne_ex}. Moreover, the representation attributes are quantified using the following metrics:
\begin{itemize}
	\item\textbf{Intra-class Variance (Variance)} (var)~\cite{hastie2009elements}: Measures class compactness by calculating the average squared distance from each point to the class centroid. A significantly lower variance than in the retrained model suggests that the unlearned data are densely clustered, indicating knowledge residue. 
	\item\textbf{Silhouette Score ($s$)}~\cite{shahapure2020cluster}: Assesses clustering quality through the average silhouette coefficient for each point within the unlearned class, ranging from -1 to 1. High scores suggest that the model retains familiarity with class knowledge, pointing to knowledge residue. 
	\item\textbf{Overlap Degree (Overlap) ($o$)}~\cite{anderson2012nonparametric}: Evaluates the overlap between the unlearned subset and others using kernel density estimation (KDE). A lower score suggests an effective classification of unlearned data, indicating knowledge residue.
\end{itemize}

\subsection{Supplementary Description of Setups and Experiments in Concept Unlearning}
\label{subsec:app_gen}
\subsubsection{Setups}  

\noindent\textbf{Models}. 
We utilize the pretrained stable diffusion model \texttt{stable-diffusion-v1-5}~\cite{rombach2022high} with prior preservation enabled, fine-tuning both the text encoder and diffusion model for 50 epochs using AdamW optimizer (learning rate is $5$e$^{-7}$), consistent with the work~\cite{van2023anti}. 

\noindent\textbf{OUR Setups}.
For OUR unlearning method, the orthogonal unlearning is applied only to output layers of the last two hidden modular blocks (\texttt{up\_blocks}) and the final convolutional layer (\texttt{conv\_out}), with scaled orthogonal loss factors [$5$e$^{-9}$, $5$e$^{-8}$] to stabilize training. The two-phase optimization spans $10$ epochs: $1$ epoch for orthogonal unlearning (learning rate is $5$e$^{-6}$) and $9$ epochs for the replay phase (learning rate is $8$e$^{-7}$), both using AdamW. Since the generation task involves fine-tuning a highly parameterized model, we do not apply L1 regularization to avoid compromising its capability.

\noindent\textbf{Concept-Unlearning Benchmark Setups}.
SalUn~\cite{fan2024salun} utilizes relabeled prompts (e.g., ``This is a photo of James'') to erase targeted concepts, while Meta Unlearning~\cite{gao2024meta} hinders the reconstruction of unlearned concepts from related concepts during unlearning, which is described by ``this is a photo of a woman''. In terms of parameter configuration, SalUn and SFRon adopt identical hyper-parameters with learning rates of $1$e$^{-6}$ across $10$ training epochs, whereas Meta Unlearning operates at a learning rate of $1$e$^{-5}$ over $50$ optimization steps. These settings strictly adhere to their original designs.

\noindent\textbf{Privacy Attack Setups}. For ReA, distinct resonance indexes emerge between the unlearned and out-of-distribution (OOD) classes when fine-tuning with only the inferred class. Thus, ReA does not rely on retained or pseudo-retained datasets in concept unlearning. It employs a learning rate of $5$e$^{-6}$, and each inferred class consists of four samples. The prompt it used is a random name, e.g., Emily. For DiffAtk~\cite{zhang2024generate}, the number of adversarial tokens is set to $5$, and the embedding method for the adversarial prompt is prefix-based.

\subsubsection{Metrics}
The evaluation metrics for generated portrait outputs (defined in Section~\ref{subsec:ex_gen}) are detailed as follows.
\begin{itemize}
	\item \textbf{Identity Score Matching (ISM)}~\cite{van2023anti}. It evaluates identity consistency between generated and reference faces using ArcFace embeddings~\cite{deng2019arcface}. Lower values indicate better identity preservation. For identity-specific concept unlearning, we introduce two metrics: \textbf{Unlearning ISM (UISM)}, which evaluates the ISM of the unlearned identity, and \textbf{Retaining ISM (RISM)}, which evaluates the ISM of the retained identity. Both are used to observe how effectively the unlearned identity is removed and the remaining identity is preserved.
	
	\item \textbf{Fréchet Inception Distance (FID)}~\cite{kabra2024f}. It quantifies the similarity between real and generated face distributions via Inception-v3 features. Lower scores reflect closer alignment to real data statistics.  
	
	\item \textbf{BRISQUE}~\cite{mittal2012no}. It assesses perceptual quality without reference by detecting unnatural patterns (e.g., artifacts, textures) in spatial features. Lower values denote more natural outputs.  
	
	\item \textbf{Resonance Index in ReA Attack} (Idx\textsubscript{r}). It measures the speed of identity recovery during the ReA process, which records the iteration where the ISM reaches $0.7$.
	
\end{itemize}

\begin{table}
	\caption{Performance of Identity-specific Concept Unlearning}
	\vspace{-0.15cm}
	\label{tb:generation_mu_sfron}
	\centering
	\tiny
	\begin{spacing}{1.0}
		\begin{tabular}{p{0.75cm}<{\centering}|p{0.95cm}<{\centering}p{0.95cm}<{\centering}p{0.9cm}<{\centering}p{0.86cm}<{\centering}p{0.85cm}<{\centering}}
			\hline
			Metrics&UISM&RISM&FID(\scalebox{0.8}{$\downarrow$})&BRISQUE(\scalebox{0.8}{$\downarrow$})&RTE (s)\\ \hline
			Retrain&0.13\scalebox{0.7}{$\pm$0.02}(\textcolor{blue}{0.00})&0.76\scalebox{0.7}{$\pm$0.02}(\textcolor{blue}{0.00})&101.82\scalebox{0.7}{$\pm$8.33}&10.76\scalebox{0.7}{$\pm$1.38}&937.50\\
			SFRon~\cite{huang2024unified}&0.50\scalebox{0.7}{$\pm$0.08}(\textcolor{blue}{0.37})&0.59\scalebox{0.7}{$\pm$0.03}(\textcolor{blue}{0.17})&134.93&17.58&483.15\\ \hline
		\end{tabular}	
	\end{spacing}	
\end{table}

\subsubsection{Experimental Results on an Additional Concept Unlearning Benchmark: SFRon~\cite{huang2024unified}}  

This section presents an evaluation of SFRon~\cite{huang2024unified} as an additional concept unlearning benchmark. Table~\ref{tb:generation_mu_sfron} shows its performance in identity-specific concept unlearning, where UISM remains as high as $0.50$, indicating ineffective removal of identity information. Table~\ref{tb:image_mu_sfron} provides further evidence, showing that the unlearning process mainly reduces image quality while preserving key identity features. As a result, SFRon is very vulnerable when attacked by ReA. It converges quickly in ReA, as shown in Table~\ref{tb:rea_generation_sfron}, where its Resonance Index reaches only $10$, far below the retrained model’s $70$.

\begin{table}
	\caption{Privacy Attack against Identity-specific Concept Unlearning}
	\vspace{-0.1cm}
	\centering
	\label{tb:rea_generation_sfron}
	\tiny
	\begin{spacing}{1.0}
		\begin{tabular}{|p{1.25cm}<{\centering}|p{2.55cm}<{\centering}p{2.55cm}<{\centering}|}
			\hline
			\multirow{2}{*}{Attacks}&\multicolumn{2}{c|}{Resonance Index, Idx\textsubscript{r}, (Maximum = 10) \scalebox{0.85}{$\times$ 10} @ ISM} \\ \cline{2-3}
			&Retrain&SFRon~\cite{huang2024unified}\\ \hline
			DiffAtk~\cite{zhang2024generate}&10\scalebox{0.7}{$\pm$0.0}@0.18&10\scalebox{0.7}{$\pm$0.0}@-0.04\\ \hline
			ReA&6.5\scalebox{0.7}{$\pm$1.0}@0.71&4\scalebox{0.7}{$\pm$0.5}@0.79\\ \hline
		\end{tabular}
	\end{spacing}
\end{table}

\begin{table}
	\caption{Images of Identity-specific Concept Unlearning}
	\vspace{-0.1cm}
	\centering
	\label{tb:image_mu_sfron}
	\tiny
	\begin{tabular}{p{0.94cm}<{\centering} p{0.945cm}<{\centering} p{0.945cm}<{\centering}}
		\hline
		Target&Retrain&SFRon~\cite{huang2024unified}\\
		\hline
		\multicolumn{3}{c}{Unlearned Identity. Prompt: ``{{\fontfamily{pzc}\selectfont a photo of Sarah}}"}\\ 
		\includegraphics[width=1.21cm]{unlearn_sarah_5_target-1.jpg}&\includegraphics[width=1.21cm]{unlearn_sarah_retrain.jpg}&\includegraphics[width=1.21cm]{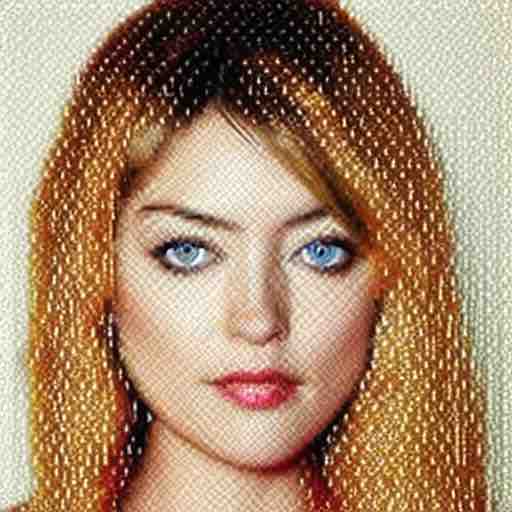}\\ \hline
		\multicolumn{3}{c}{Retained Identity. Prompt: ``{\fontfamily{pzc}\selectfont \textbf{a photo of Laura}}"}\\ 
		\includegraphics[width=1.21cm]{2127.jpg}&\includegraphics[width=1.21cm]{retain_laura_retrain.jpg}&\includegraphics[width=1.21cm]{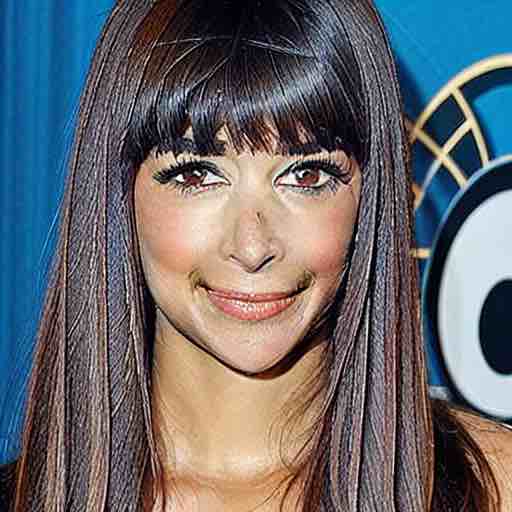}\\ \hline
	\end{tabular}
\end{table}

\begin{figure}
	\centering
	\subfloat[FT]{
		\includegraphics[width=0.4\columnwidth]{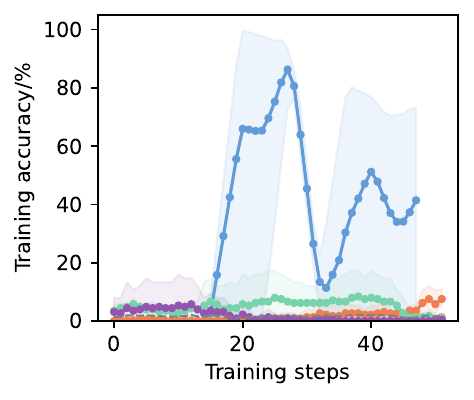}}
	\subfloat[GA]{
		\includegraphics[width=0.4\columnwidth]{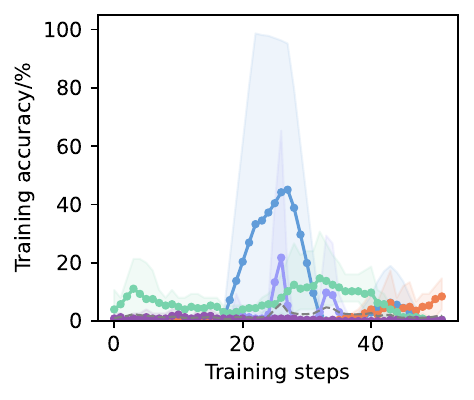}}\\
	\makebox[0.04cm]{}
	\subfloat[RL]{
		\includegraphics[width=0.4\columnwidth]{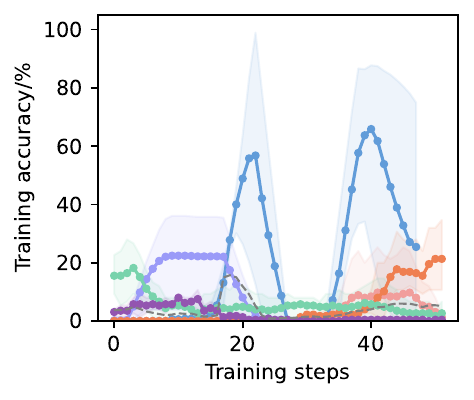}}
	\subfloat[IU (WoodFisher)]{
		\includegraphics[width=0.4\columnwidth]{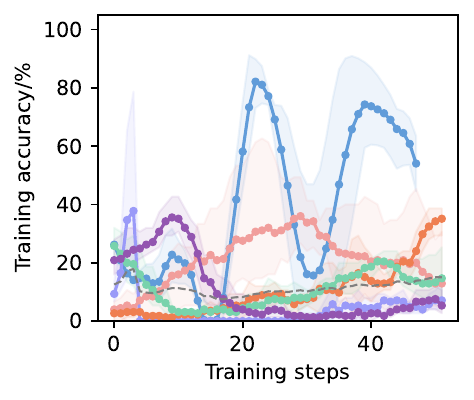}}
	\makebox[0.31cm]{}
	\includegraphics[width=0.88\columnwidth]{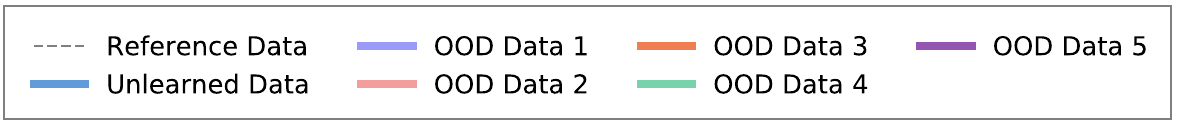}
	\caption{Training trajectories in the \emph{Reminiscence Attack} under black-box scenarios. $3\%$ of each type of data is available. The light-colored areas represent the error bars at different learning rates.}
	\label{fig:training_trajectory_bk}
\end{figure}

\subsection{Evaluating Reminiscence Attack under Black-box Scenarios}
\label{subsec:scenario}
We investigate the class-wise ReA attack performance in black-box scenarios, to explore the transferability of knowledge residue from approximate unlearning in model extraction~\cite{truong2021data}. In white-box attacks, the attacker has direct access to the target model, allowing the ReA to be launched immediately. Conversely, in black-box scenarios, an additional preparatory step is required. This involves using data-free model extraction (ME) attacks (DFME)~\cite{truong2021data} to obtain a local substitute model on which the ReA is then launched. In black-box scenarios, we set the attacker's query budget to $20$M and $30$M in the ME step for Cifar20 (-5) and Cifar100 (-5), around the primary setup of DFME~\cite{truong2021data}. Four approximate unlearning benchmarks are evaluated.

Table~\ref{tb:fidelity} presents the accuracy and fidelity of substitute models obtained via DFME, and Table~\ref{tb:scenario} reports the attack performance and computational cost of class-wise ReA against different unlearning methods. The adversarial set size is $3\%$ for CIFAR20 (-5) and $20\%$ for CIFAR100 (-5). Additionally, Figure~\ref{fig:training_trajectory_bk} visualizes the reminiscence process in black-box settings, revealing that despite instability in training due to fidelity loss in ME, the unlearned dataset consistently exhibits distinct resonance.

\noindent\textbf{Results}. Overall, ReA remains highly effective in black-box scenarios, nearly matching its white-box performance despite higher computational costs. This suggests that \textbf{local models from ME inherit significant class-type residual knowledge}, exposing fundamental vulnerabilities in approximate unlearning methods.

\begin{table}
	\caption{Model Extraction Performance}
	\label{tb:fidelity}
	\tiny
	\begin{spacing}{1.1}
		\begin{tabular}{|p{1.6cm}<{\centering}|p{1.13cm}<{\centering}|p{1.11cm}<{\centering}|p{1.11cm}<{\centering}|p{1.11cm}<{\centering}|}
			\hline
			\multirow{2}{*}{Dataset}& \multicolumn{4}{c|}{ME Accuracy (Fidelity) / $\%$} \\ \cline{2-5}
			& FT&GA&RL&IU\\ \hline
			\multirow{2}{*}{Cifar20 (-5)}&60.36&61.64&58.81&56.02 \\
			&(77.73)&(78.79)& (73.65)&(69.53)\\ \hline
			\multirow{2}{*}{Cifar100 (-5)}&54.83&56.74&58.81&56.29\\
			&(69.14)&(71.24)&(73.63)&(70.54)\\ \hline
		\end{tabular}
	\end{spacing}
\end{table}

\begin{table}
	\caption{\emph{Reminiscence Attack} Under Black-Box Scenarios}
	\label{tb:scenario}
	\tiny
	\begin{spacing}{1.1}
		\begin{tabular}{|p{1.1cm}<{\centering}|p{0.94cm}<{\centering}|p{0.94cm}<{\centering}|p{0.94cm}<{\centering}|p{0.94cm}<{\centering}|p{0.75cm}<{\centering}|}
			\hline
			\multirow{2}{*}{Dataset}&\multicolumn{4}{c|}{Accuracy (TPR @ $0.1$ FPR) / $\%$}&Cost \\ \cline{2-5}
			&FT& GA&RL& IU&(s)\\ \hline
			\multirow{2}{*}{Cifar20 (-5)}&83.00&65.50&82.50&77.50&\multirow{2}{*}{9.16e$^3$}\\ 
			&(90.00)&(55.00)&(70.00)&(80.00)&\\ \hline
			\multirow{2}{*}{Cifar100 (-5)}&80.00&58.50&76.25&80.25&\multirow{2}{*}{1.36e$^4$}\\ 
			&(85.00)&(40.00)&(57.50)&(65.00)&\\ \hline
		\end{tabular}
	\end{spacing}
\end{table}

\end{document}